\documentclass[a4paper,fleqn]{cas-dc}

\usepackage{natbib} 
\usepackage{graphicx}
\usepackage{caption}
\usepackage{subcaption}
\usepackage{multirow}
\usepackage{amsmath} 
\usepackage{makecell}
\usepackage{hyperref}
\usepackage{float}
\usepackage[switch]{lineno}

\newcommand\ParticleSizeMM{\textit{ParticleSize mm}}
\newcommand\ParticleSizeVoxel{\textit{ParticleSize voxel}}
\newcommand\VoxelSpacingMM{\textit{VoxelSpacing mm}}

\makeatletter
\newcommand{\phantomlabel}[2]{
    \protected@write\@auxout{}{
        \string\newlabel{#2}{
            {\@currentlabel#1}{\thepage}
            {\@currentlabel#1}{#2}{}
        }
    }
    \hypertarget{#2}{}
}
\makeatother

\begin{document}

\let\WriteBookmarks\relax
\def\floatpagepagefraction{1}
\def\textpagefraction{.001}

\shorttitle{ParticleSeg3D: A Scalable Out-of-the-Box Deep Learning Segmentation Solution for Individual Particle Characterization from Micro CT Images in Mineral Processing and Recycling}

\shortauthors{Karol Gotkowski et~al.}

\title [mode = title]{ParticleSeg3D: A Scalable Out-of-the-Box Deep Learning Segmentation Solution for Individual Particle Characterization from Micro CT Images in Mineral Processing and Recycling}                      

\author[1, 2]{Karol Gotkowski}

\ead{karol.gotkowski@dkfz.de}

\author[3]{Shuvam Gupta}
\ead{s.gupta@hzdr.de}

\author[3]{Jose R. A. Godinho}
\ead{j.godinho@hzdr.de}

\author[3]{Camila G. S. Tochtrop}
\ead{c.guimaraes-da-silva-tochtrop@hzdr.de}

\author[2, 4]{Klaus H. Maier-Hein}
\ead{klaus.maier-hein@dkfz-heidelberg.de}

\author[1, 2]{Fabian Isensee}
\ead{f.isensee@dkfz-heidelberg.de}

\cormark[1]

\affiliation[1]{organization={Helmholtz Imaging, German Cancer Research Center},
    addressline={Im Neuenheimer Feld 280}, 
    city={Heidelberg},
    postcode={69120},  
    state={Baden-Wuerttemberg},
    country={Germany}
    }

\affiliation[2]{organization={Division of Medical Image Computing, German Cancer Research Center},
    addressline={Im Neuenheimer Feld 280}, 
    city={Heidelberg},
    postcode={69120}, 
    state={Baden-Wuerttemberg},
    country={Germany}
    }

\affiliation[3]{organization={Helmholtz-Zentrum Dresden-Rossendorf, Helmholtz Institute Freiberg for Resource Technology},
    addressline={Chemnitzer Straße 40}, 
    city={Freiberg},
    postcode={09599}, 
    state={Sachsen},
    country={Germany}}

\affiliation[4]{organization={Pattern Analysis and Learning Group, Department of Radiation Oncology, Heidelberg University Hospital},
    addressline={Im Neuenheimer Feld 672}, 
    city={Heidelberg},
    postcode={69120}, 
    state={Baden-Wuerttemberg},
    country={Germany}}

\cortext[cor1]{Corresponding author}


\begin{abstract}
Minerals, metals, and plastics are indispensable for a functioning modern society. Yet, their supply is limited causing a need for optimizing ore extraction and recuperation from recyclable materials. Typically, those processes must be meticulously adapted to the precise properties of the processed materials. Advancing our understanding of these materials is thus vital and can be achieved by crushing them into particles of micrometer size followed by their characterization. Current imaging approaches perform this analysis based on segmentation and characterization of particles imaged with computed tomography (CT), and rely on rudimentary postprocessing techniques to separate touching particles. However, their inability to reliably perform this separation as well as the need to retrain methods for each new image, these approaches leave untapped potential to be leveraged. Here, we propose ParticleSeg3D, an instance segmentation method able to extract individual particles from large CT images of particle samples containing different materials. Our approach is based on the powerful nnU-Net framework, introduces a particle size normalization, uses a border-core representation to enable instance segmentation, and is trained with a large dataset containing particles of numerous different sizes, shapes, and compositions of various materials. We demonstrate that ParticleSeg3D can be applied out-of-the-box to a large variety of particle types, including materials and appearances that have not been part of the training set. Thus, no further manual annotations and retraining are required when applying the method to new particle samples, enabling substantially higher scalability of experiments than existing methods. Our code and dataset are made publicly available.
\end{abstract}


\maketitle

\section{Introduction}\label{Introduction}

Material resources such as minerals, metals, and plastics are ubiquitous in modern society, yet, they constitute a finite resource that needs to be acquired cost-effectively and used responsibly. Moreover, with the ever-rising demand for specific materials, these processing and recycling processes are becoming increasingly automated. Characterization of samples containing crushed and pulverized particles of micrometer size from these materials plays a crucial role in this process by providing a detailed understanding of the material composition alongside other factors, contributing to the development of effective processes to both mine and recycle specific materials. \\
So far, sample-level analysis of the materials present in a sample can be conducted through a multitude of methods such as X-ray diffraction, inductively coupled plasma mass spectrometry (ICP-MS), or computed tomography (CT). In the latter, the particles are embedded in an epoxy matrix and imaged via CT in order to create a semantic segmentation that segments all particles, distinguishing between different particle material types but not between particle instances. The segmentation is either performed through a variant of intensity thresholding \cite{hassan2012nondestructive, becker2016x, dominy2011characterisation, godinho2019volume} or deep learning approaches \cite{wang2021improved, filippo2021deep, xiao2020ore, latif2022deep, nie2022image, liu2021efficient, tung2022deep}. The individual particle properties that determine the behavior of each particle are not accounted for in sample-level analyses, limiting our understanding of the used materials and our ability to optimize downstream processes. \\
A better optimization can be achieved through particle-level analysis with the characteristics of individual particles \cite{pereira2021self}. Naturally, this approach is only possible with high-resolution imaging techniques such as CT. Here, every particle instance is segmented and assigned a unique identifier, resulting in a so-called instance segmentation. This has the advantage that each particle can be further characterized individually according to its specific shape, size, and particle histogram \cite{godinho20233d}. Such attempts to characterize individual particles have been mostly limited to imaging techniques that analyze only a single 2D image \cite{liu2020ore, baraian2022computing, sun2022efficient}, leading to an inherent stereological bias that makes the shape and size characterization of particles unreliable \cite{blannin2021uncertainties}. \\
Extending individual particle characterization to 3D is challenging, given that upwards of ten thousand particles are typically present in a single image. Even more, the presence of imaging artifacts as well as highly diverse materials and filler materials make this a difficult problem where established methods that rely on intensity thresholding \cite{zhou2021three, jiang2021characterisation, wang20163d, wang2015improved, guntoro2021development} may fail due to their lack of robustness. To solve this problem for 3D instance segmentation more sophisticated deep learning methods are a necessity \cite{furat2023multidimensional, tang2023particle} as they use a multitude of learned features, making them inherently more robust to these challenges. However, a limitation of these approaches is constituted through the use of semantic segmentation models to generate predictions that are converted in a separate postprocessing step into instance segmentations, leading to touching particles often not recognized as separate instances and thus hindering individual particle analysis. Moreover, a common drawback of all current deep learning approaches used for particle segmentation is the narrow scope on specific types of materials, resulting in a lack of generalization. Consequently, these methods do not work out-of-the-box when applied to a particle image containing new materials as part of the image needs to be annotated manually and the model retrained in order to function properly. \\
To truly pave the way for process optimization based on individual 3D particle characterization, we propose ParticleSeg3D, an automated out-of-the-box 3D instance segmentation method with high robustness and generalizability that can handle diverse sample and particle types across magnitudes of different particle sizes without the need to adapt the method or annotate new data and thus does not constitute major bottlenecks. Our approach employs a highly optimized novel training and inference pipeline based on a 3D U-Net and is trained on a large dataset of CT particle samples with a multitude of different materials such as various natural ores, slags with complex microstructures, batteries, and other electronics with plastics and large variations of components. Thus our approach enables a wide range of analyses based on the intrinsic particle properties measurable based on the inferred instance segmentations and can be employed even by non-machine learning experts without prior knowledge.  
Our framework and all data are open source and available under \href{https://github.com/MIC-DKFZ/ParticleSeg3D}{https://github.com/MIC-DKFZ/ParticleSeg3D}.

\section{Materials}\label{Materials}

\subsection{Sample preparation}\label{Sample preparation}

The samples used in our approach consist of various materials crushed and pulverized into particles of micrometer sizes. They were prepared following a standardized procedure described in \cite{godinho2021mounted} which uses sugar particles as spacers in the ratio of 7 g of sugar to 1 g of particles. The resin used is Paladur (Kulzer, Mitsui Chemical Group), a fast-curing acrylic polymer. The material particles together with the sugar were mixed with methylmethacrylate-copolymer powder in a mass ratio of 1:1. The methylmethacrylate liquid resin was added to the solid mixture in a ratio of 3 mL to 10 g. The final paste is left to dry in a tube with a diameter adequate for the desired voxel spacing of the scan. \\
Samples were imaged with a CT scanner (CoreTom from XRE – Tescan; Ghent, Belgium), while the  “XRE – recon” software (v1.1.0.14, XRE – Tescan, Ghent, Belgium) was used to reconstruct the 3D images in 16-bit. Scanning conditions differed on a per-sample basis as to use optimal reconstruction parameters based on material compositions. Consequently, voxel intensities and particle sizes vary between the obtained images.

\subsection{Datasets} \label{Datasets}

\begin{figure*}
\centering
        \begin{subfigure}[b]{0.3\textwidth}  
                \centering
                \includegraphics[width=.85\linewidth]{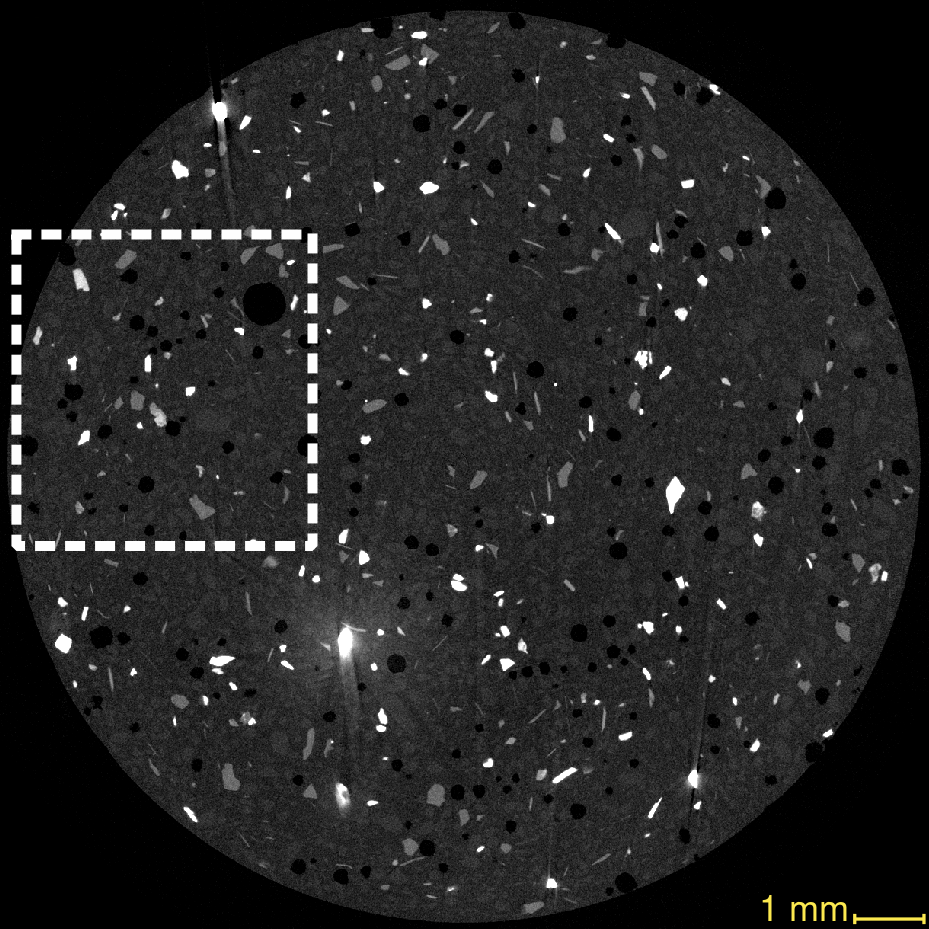}
                \caption{Sample}
        \end{subfigure}%
        \begin{subfigure}[b]{0.3\textwidth}
                \centering
                \includegraphics[width=.85\linewidth]{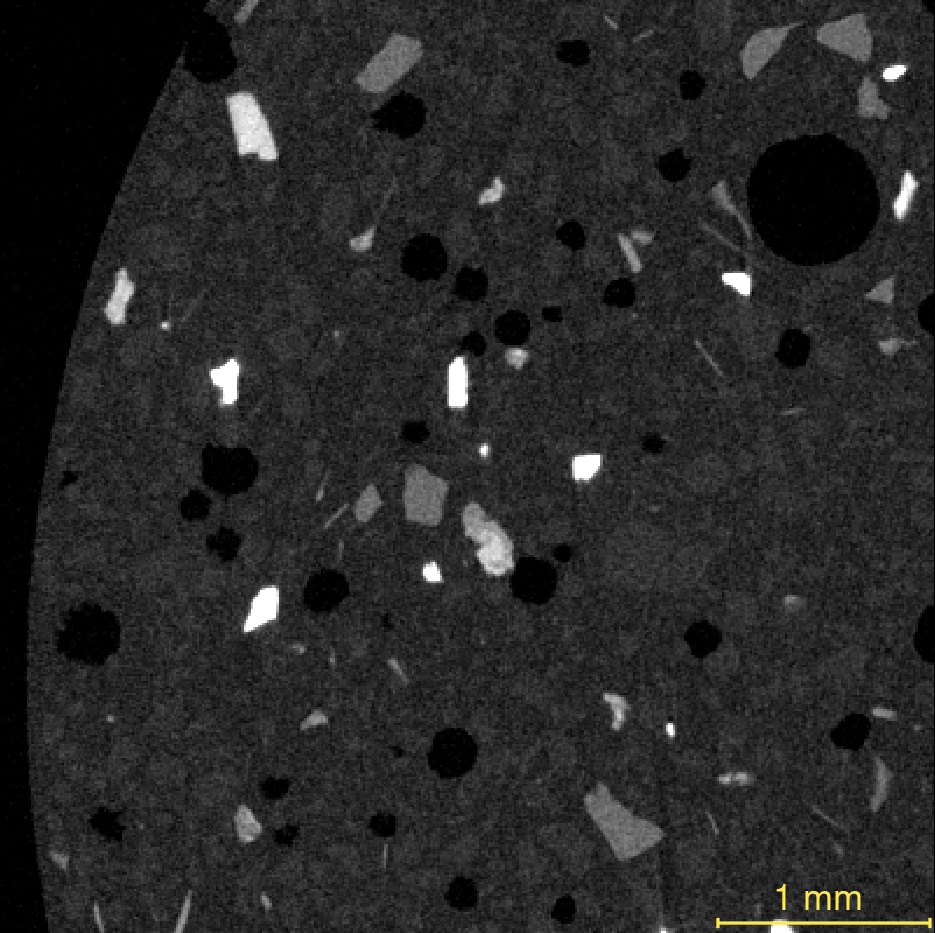}
                \caption{Sample patch}
        \end{subfigure}%
        \begin{subfigure}[b]{0.3\textwidth}
                \centering
                \includegraphics[width=.85\linewidth]{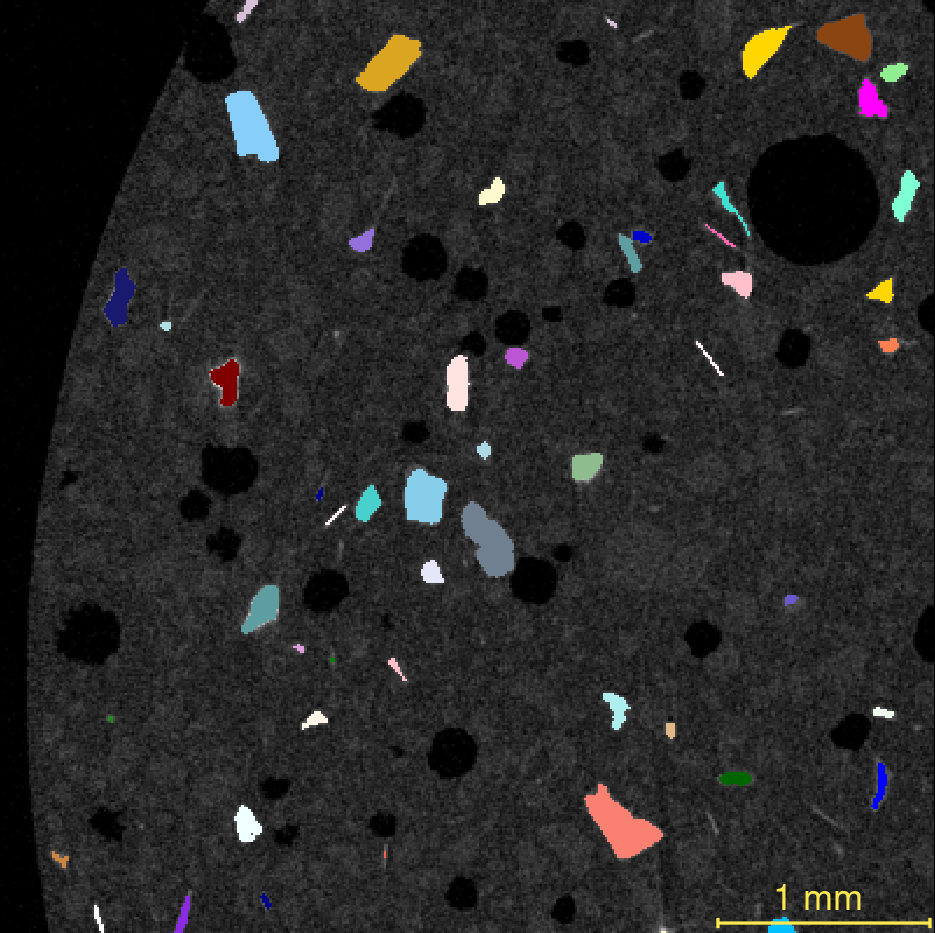}
                \caption{Patch reference segmentation}
        \end{subfigure}%
        \caption{Overview of the image Synthetic1 and its reference segmentation.}
        \label{fig:datasets:overview}
\end{figure*}

\begin{figure}
    \centering
    \includegraphics[width=0.9\linewidth]{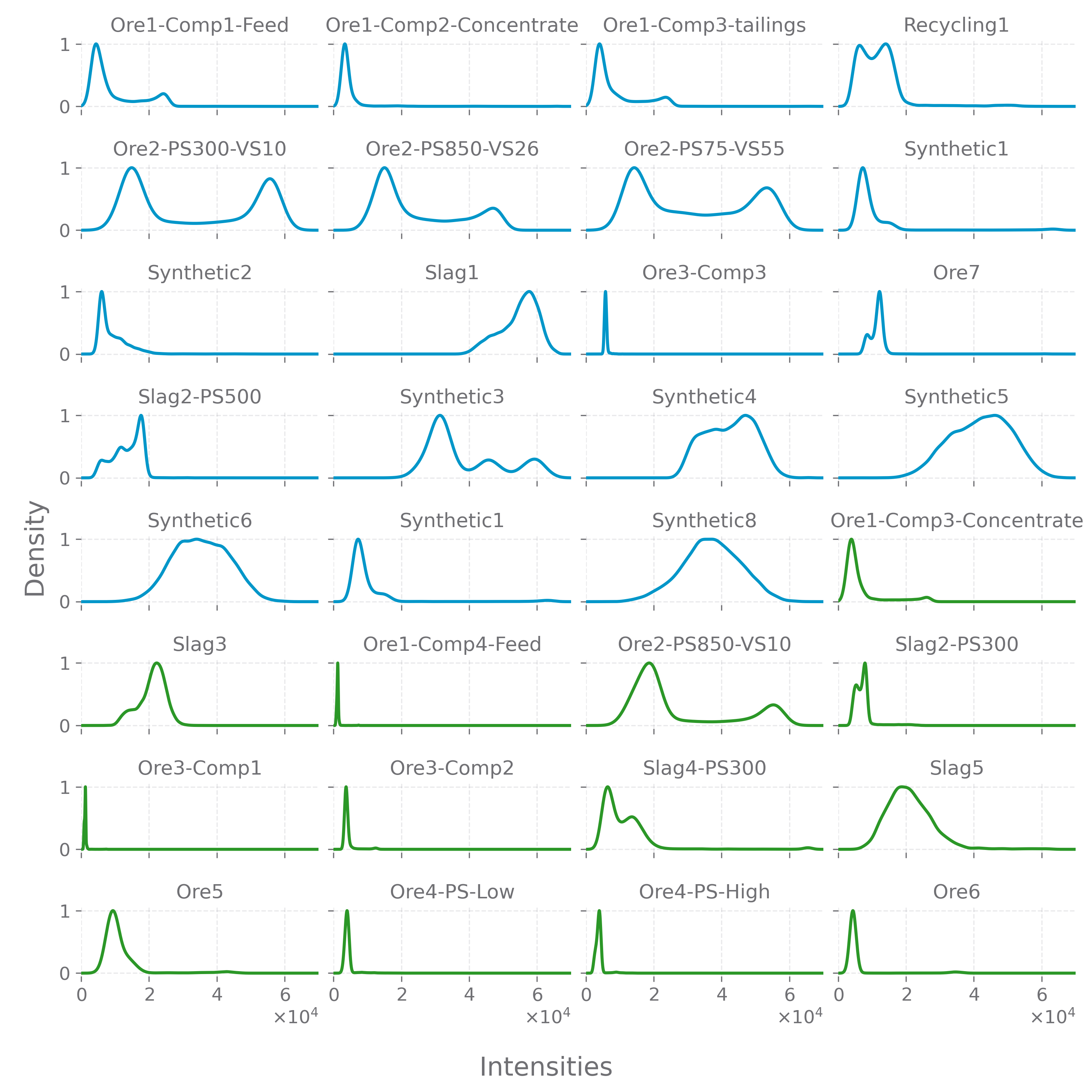}
    \caption{Kernel Density Estimation of the different particle intensity distributions of the train set (blue) and the test sets (green). Densities have been normalized to [0,1] for each sample.}
    \label{fig:datasets:intensities}
\end{figure}

A diverse set of samples is required in order to train a model that predicts high-accuracy instance segmentations and generalizes well even to materials not present in the training distribution. The samples prepared for our approach cover a wide range of materials, e.g. various natural ores, slags with complex microstructures, batteries, and other electronics with plastics and large variations of components to fulfill this criterion. The mean image size of a sample is (997±287, 1256±218, 1256±218) voxels and the mean particle size of different samples ranges from 55 to 721 micrometers as determined through the equivalent particle diameter but can also be substantially smaller and larger due to strong variance in particle sizes within samples. For the development and training of our method, 19 samples were prepared in total from which we used 41 patches with either a size of 128x400x400 or 200x400x400 voxels that had been extracted and annotated following the procedure outlined in section \ref{Reference annotation}. An example of such a prepared and annotated sample is shown in Figure \ref{fig:datasets:overview}. Further, we depict in Figure \ref{fig:datasets:intensities} the particle intensity distributions of all samples, with the intensity peaks corresponding to different materials exhibiting the diverse nature of our samples. \\
For testing the performance of our model, we designed three test sets reflecting different use cases. The first is an in-distribution set of in total 8 patches taken from 8 unseen samples, i.e. samples that are not part of the train set, with the purpose to evaluate the performance in terms of its prediction accuracy on samples that consist of materials already seen by the framework during training. The second is an out-of-distribution set with 5 patches taken from 5 samples that are not part of the train set and that consist of materials entirely or in their composition unknown to the model to evaluate the generalization abilities of our method. The third is a set consisting of a single sample that is not part of the train set with most particles strongly affected by streak artifacts as a result of beam hardening. No reference segmentation can be reliably created with the procedure in section \ref{Reference annotation} and thus this sample is only used for the qualitative evaluation in section \ref{Qualitative evaluation} to test the robustness of our method against artifacts. A listing of all samples is given in Table \ref{table:materials:datasets:sample_listing}.

\begin{table}[h!]
\centering
\caption{A listing of all particle samples, their primary material composition, and physical particle size range that have been used for the train set, in-distribution (ID) test set, out-of-distribution (OOD) test set, and artifacts (A) test set.}
\def\arraystretch{1.2}
\resizebox{\columnwidth}{!}{  
\begin{tabular}{| c | c | c | c |} 
\hline 
\textbf{Sample} & \textbf{Primary Materials} & \textbf{Particle Size ($\mu$m)} & \textbf{Set}\\ 
\hline\hline
Ore1-Comp1-Feed \footnotemark[1] & Sn-skarn & 122±41 & \multirow{19}{*}{Train} \\ 
\cline{1-3}
Ore1-Comp2-Concentrate \footnotemark[1] & Sn-skarn & 113±40 &  \\ 
\cline{1-3}
Ore1-Comp3-tailings \footnotemark[1] & Sn-skarn & 177±42 &  \\ 
\cline{1-3}
Recycling1 & Li-Co-Mn-Ni batteries & 156±49 & \\ 
\cline{1-3}
Ore2-PS300-VS10 & Chromite ore & 249±154 & \\ 
\cline{1-3}
Ore2-PS850-VS26 & Chromite ore & 721±304 & \\ 
\cline{1-3}
Ore2-PS75-VS55 & Chromite ore & 85±32 & \\ 
\cline{1-3}
Synthetic1 & Various ores, slags, batteries & 117±67 & \\ 
\cline{1-3}
WEEE slag & PCBs & 136±89 & \\ 
\cline{1-3}
Slag1 & Cu-Fe rich & 61±65 & \\ 
\cline{1-3}
Ore3-Comp3 \footnotemark[2] & Scheelite ore & 141±48 & \\ 
\cline{1-3}
Ore7 & Sulphide-based ore & 344±209 & \\ 
\cline{1-3}
Slag2-PS500 & Iron-rich & 471±134 & \\ 
\cline{1-3}
Synthetic2 & Quartz, calcite, fluorite & 126±98 & \\ 
\cline{1-3}
Synthetic3 & Calcite & 62±40 & \\ 
\cline{1-3}
Synthetic4 & Calcite & 55±40 & \\ 
\cline{1-3}
Synthetic5 & Lepidolite & 116±77 & \\ 
\cline{1-3}
Synthetic6 & Glass beads, calcite and fluorite & 218±92 & \\ 
\cline{1-3}
Synthetic7 & Quartz, lepidolite & 189±100 & \\ 
\hline\hline
Ore1-Comp3-Concentrate \footnotemark[1] & Sn-skarn & 133±45 & \multirow{7}{*}{ID Test} \\ 
\cline{1-3}
Slag3 & Chromium-rich & 402±150 & \\ 
\cline{1-3}
Ore1-Comp4-Feed \footnotemark[1] & Sn-skarn & 123±36 & \\ 
\cline{1-3}
Ore2-PS850-VS10 & Chromite ore & 592±252 & \\ 
\cline{1-3}
Slag2-PS300 & Iron-rich & 188±85 & \\ 
\cline{1-3}
Ore3-Comp1 \footnotemark[2] & Scheelite ore & 140±52 &  \\ 
\cline{1-3}
Ore3-Comp2 \footnotemark[2] & Scheelite ore & 49±32 &  \\ 
\cline{1-3}
Slag4-PS300 \footnotemark[3] & WEEE-recycling & 149±111 &  \\ 
\hline\hline
Slag5 & Zinc-rich & 435±134 & \multirow{5}{*}{OOD Test} \\ 
\cline{1-3}
Ore5 & Cu/Au-ore & 147±56 & \\ 
\cline{1-3}
Ore4-PS-Low \footnotemark[4] & Parisite & 85±55 & \\ 
\cline{1-3}
Ore4-PS-High \footnotemark[4] & Parisite & 59±51 &  \\ 
\cline{1-3}
Ore6 \footnotemark[5] & Sulphide concentrate & 393±141 &  \\ 
\hline\hline
Ore8 & Gold ore & 127±34 & A Test \\ 
\hline
\end{tabular}
}
\label{table:materials:datasets:sample_listing}
\end{table}

\footnotetext[1]{\cite{kern2019inherent}}
\footnotetext[2]{\cite{kern2022integration}}
\footnotetext[3]{\cite{buchmann2020evaluation}}
\footnotetext[4]{\cite{godinho20233d}}
\footnotetext[5]{\cite{winardhi2022particle}}

\newpage

\section{Methodology} \label{Methodology}

\begin{figure*}[h!]
    \centering
    \includegraphics[width=0.9\textwidth]{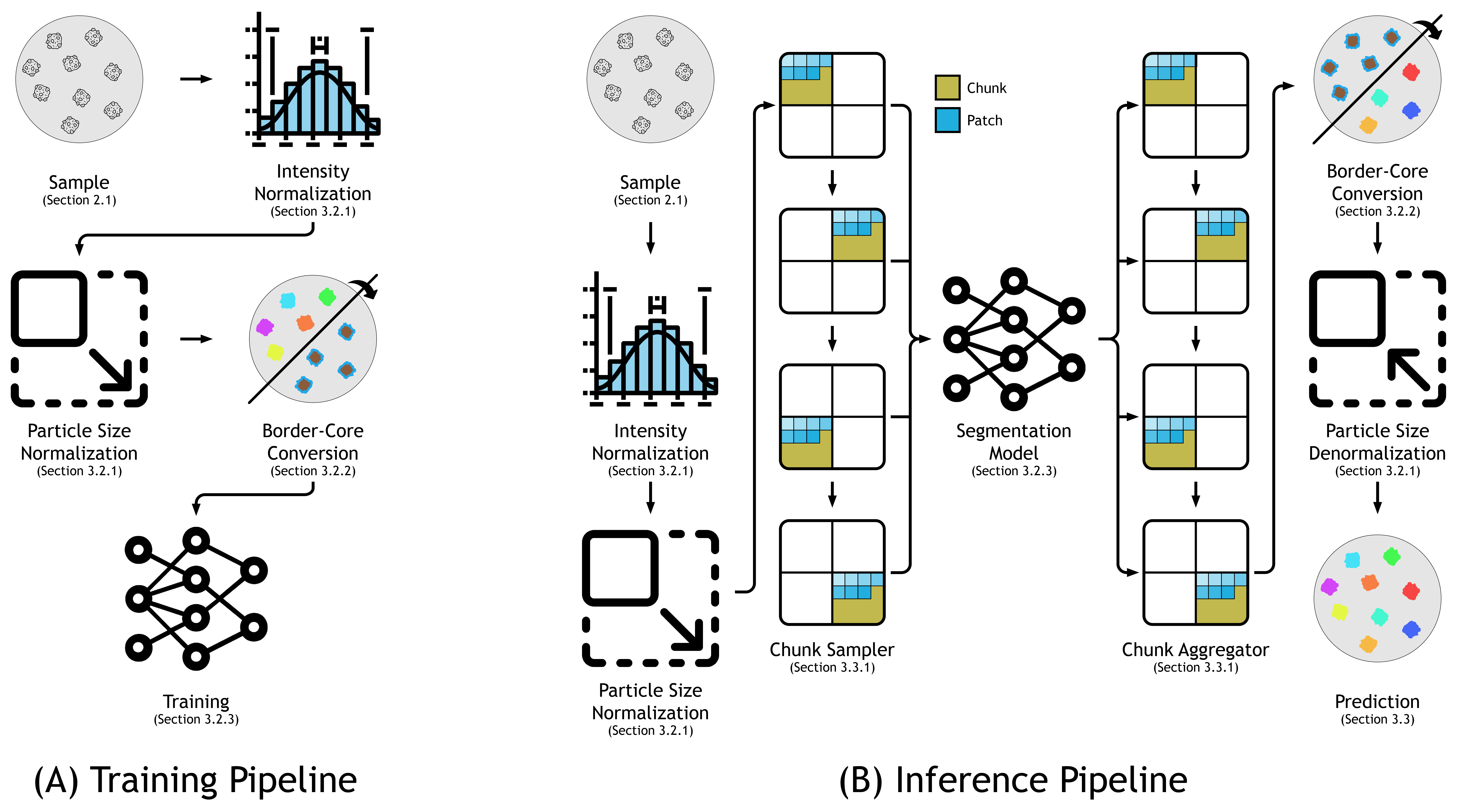}
    \caption{Training and inference pipelines of ParticleSeg3D that employ intensity and particle size normalization to increase robustness, a border-core representation to enable usage of the nnU-Net, and a chunk sampler to infer predictions for very large images.}
    \label{fig:methodology:training_and_inference_pipeline}
    \phantomlabel{.A}{fig:methodology:training_and_inference_pipeline:train}
    \phantomlabel{.B}{fig:methodology:training_and_inference_pipeline:inference}
\end{figure*}

Instance segmentation of 3D voxel images is still a problem mostly solved with conventional methods. 3D instance segmentation models based on deep learning \cite{kopelowitz2019lung, fan2020mass, zhao2018deep, zhao2021voxelembed, wang2019nodule, jeong2020brain} are so far tailored to specific, mostly medical, tasks and have not been extensively used and evaluated for generalization and robustness outside their scope. On the contrary, 3D semantic segmentation is a well-studied task with a variety of suitable models proven successful such as U-Nets and Transformers. Especially nnU-Net, a state-of-the-art 3D semantic segmentation model, has proven its performance, generalization abilities, and robustness on many medical challenges \cite{isensee2021nnu} and even non-medical applications \cite{grabowski2022self, li2022unsupervised}. However, converting the semantic segmentations predicted by such models into instance segmentations is challenging as particles cannot be separated easily due to their inconsistent shapes and tendency to touch and intertwine with each other. In order to handle such cases, we reformulate the instance segmentation problem as a semantic segmentation problem by converting the instances to a border-core representation, elaborated in \ref{Border-core representation}, enabling the usage of nnU-Net for 3D instance segmentation. \\
With this technique at its core, we propose the new deep learning based method ParticleSeg3D as depicted in Figure \ref{fig:methodology:training_and_inference_pipeline} that enables a fully automated generation of instance segmentations for individual particle characterization in 3D CT images. The optimized annotation pipeline developed for this task and proposed in section \ref{Reference annotation} is used to create training data for our method due to a lack of openly available reference instance segmentation data. Subsequently, we use the data in our training pipeline, as described in section \ref{Training pipeline}, to train a robust model that generalizes well even to unseen types of materials. As a result, no additional training data or finetuning is required for the inference process of new samples.

\subsection{Reference annotation} \label{Reference annotation}
\begin{figure}
\centering
        \begin{subfigure}[b]{0.3332\columnwidth}
                \centering
                \includegraphics[width=.95\linewidth]{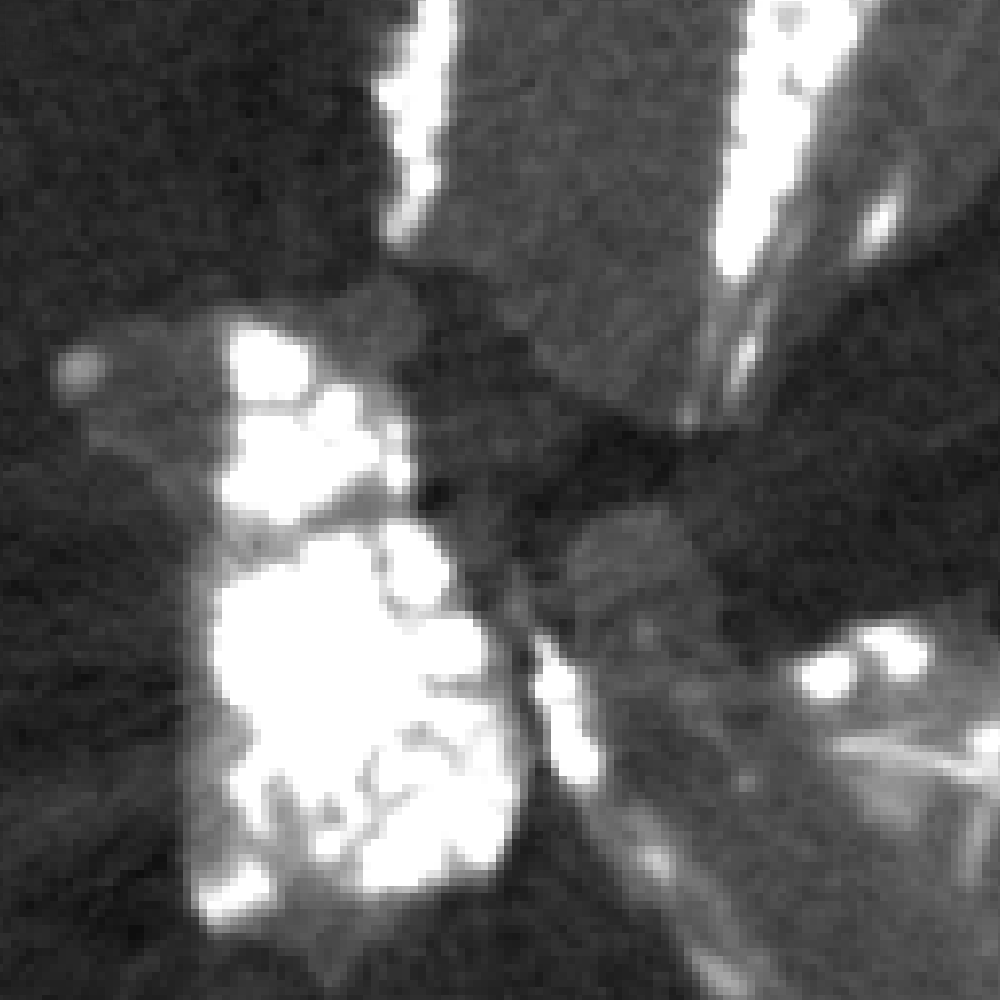}
                \caption{Patch}
        \end{subfigure}%
        \begin{subfigure}[b]{0.3332\columnwidth}
                \centering
                \includegraphics[width=.95\linewidth]{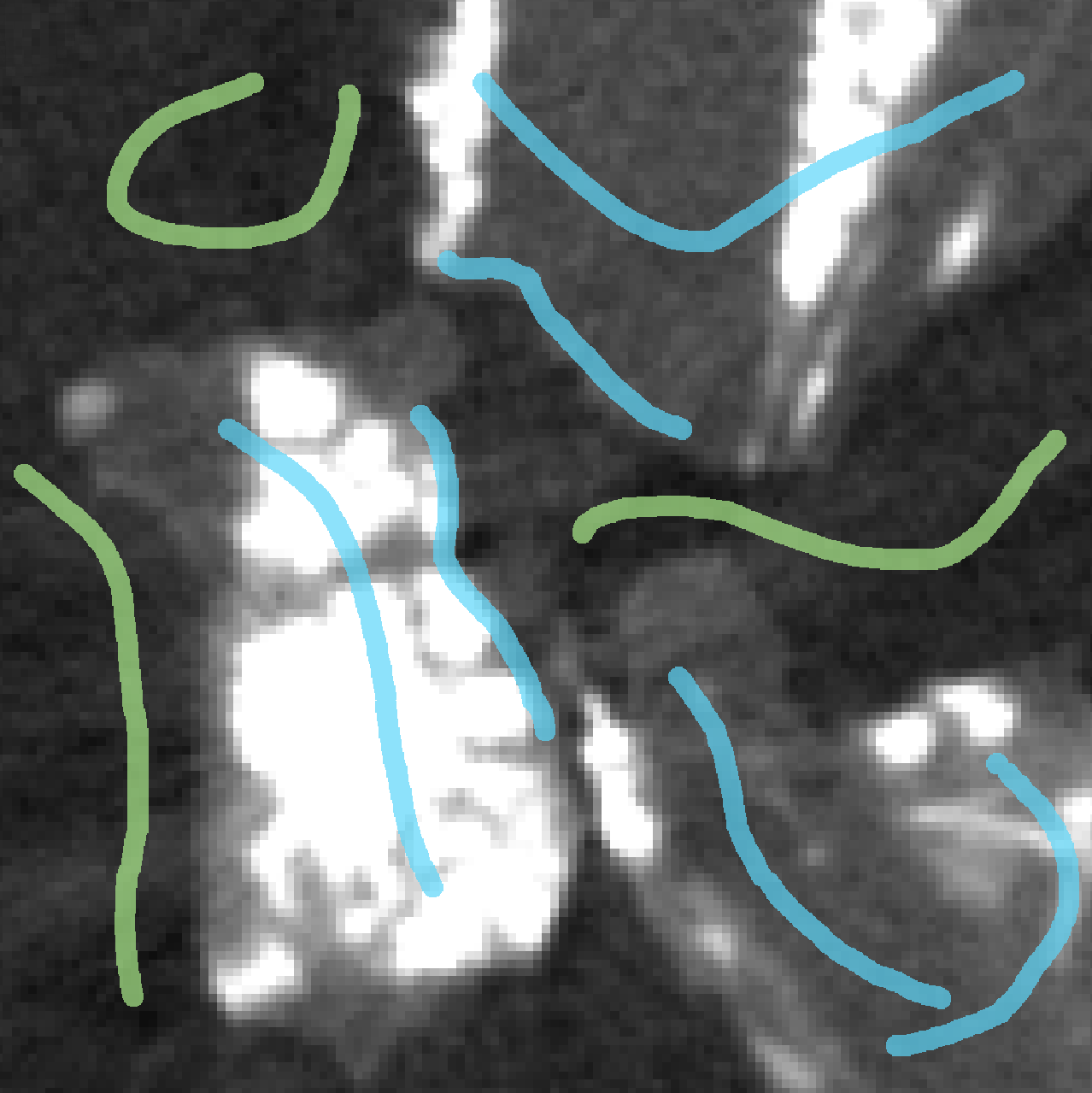}
                \caption{Scribbles}
        \end{subfigure}%
        \begin{subfigure}[b]{0.3332\columnwidth}
                \centering
                \includegraphics[width=.95\linewidth]{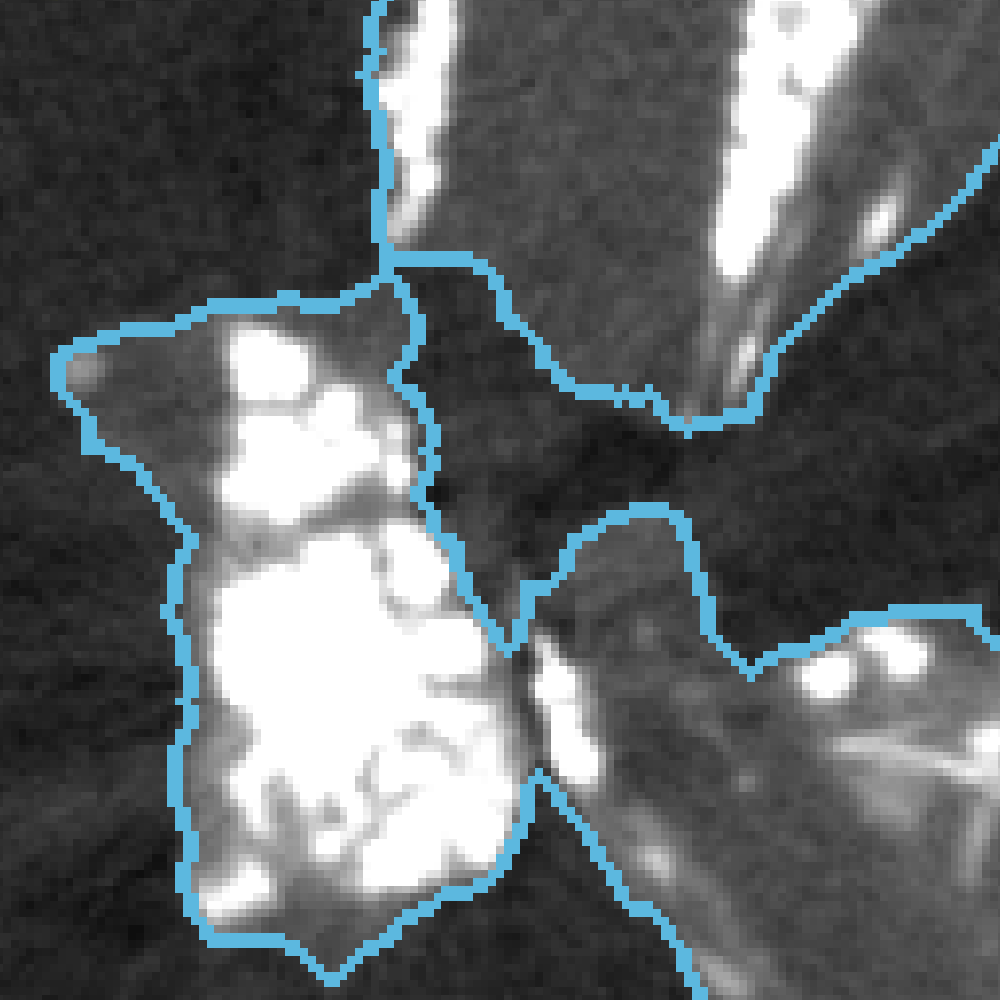}
                \caption{Segmentation}
        \end{subfigure}%
        \caption{Annotation process of creating a semantic segmentation with Random Forest through iterative refinement with scribbles. (a) A patch is extracted from the image; (b) Scribbles are iteratively refined in the patch; (c)  A semantic segmentation of the particles is successfully created.}
        \label{fig:methodology:reference_annotation}
\end{figure}

\begin{figure}
\centering
        \begin{subfigure}[b]{0.3332\columnwidth}
                \centering
                \includegraphics[width=.95\linewidth]{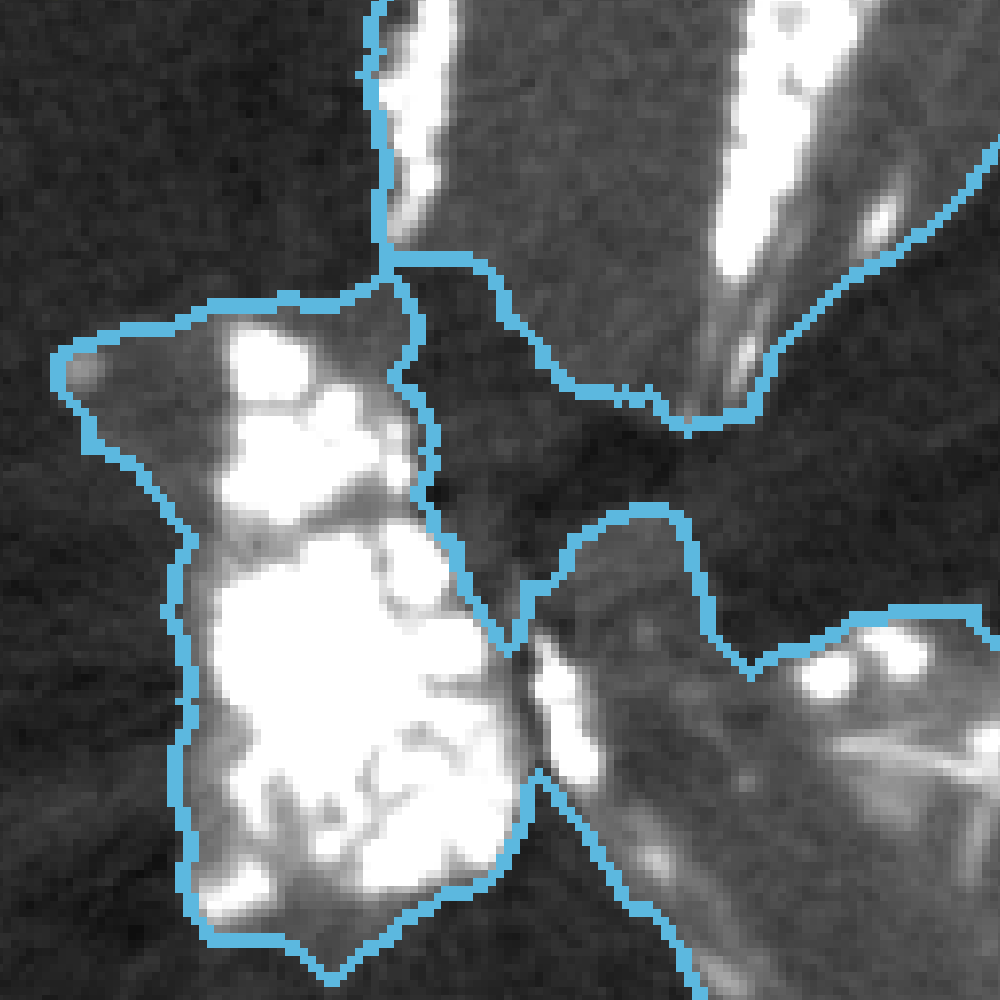}
                \caption{Touching Particles}
        \end{subfigure}%
        \begin{subfigure}[b]{0.3332\columnwidth}
                \centering
                \includegraphics[width=.95\linewidth]{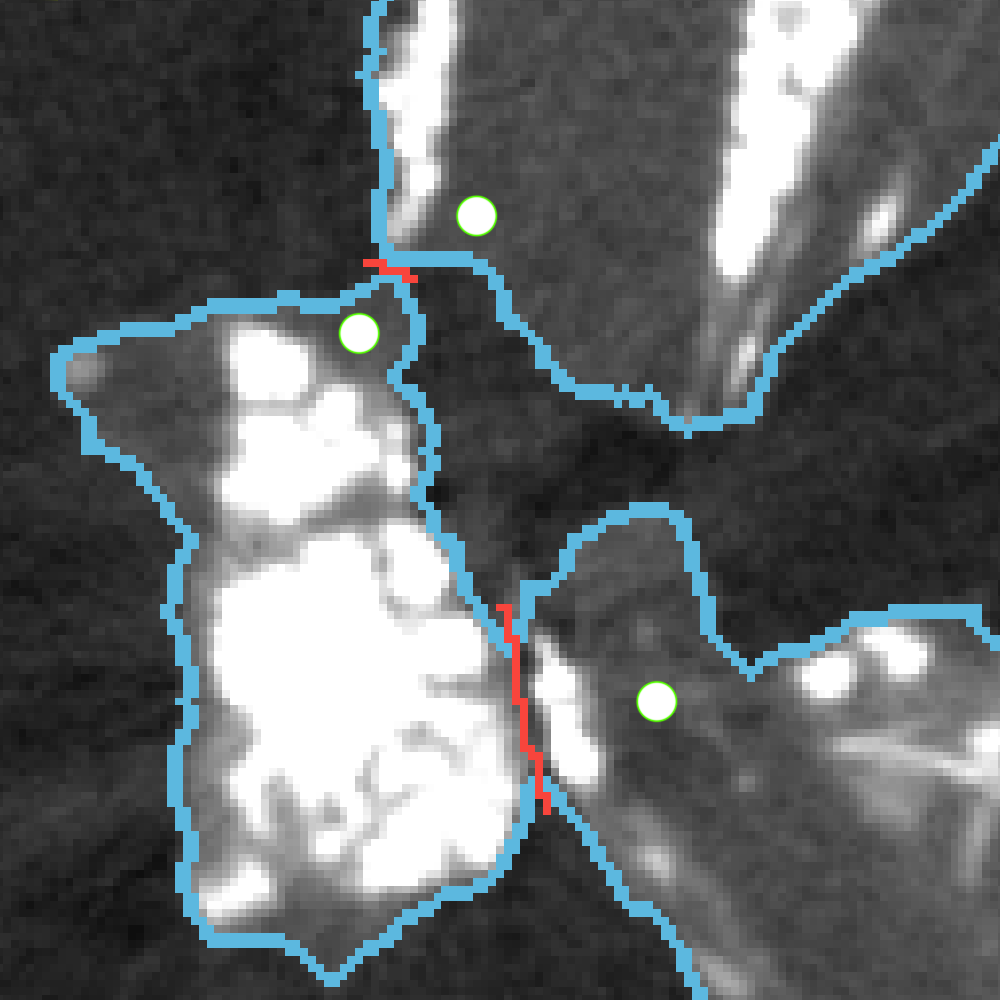}
                \caption{Splitting Process}
        \end{subfigure}%
        \begin{subfigure}[b]{0.3332\columnwidth}
                \centering
                \includegraphics[width=.95\linewidth]{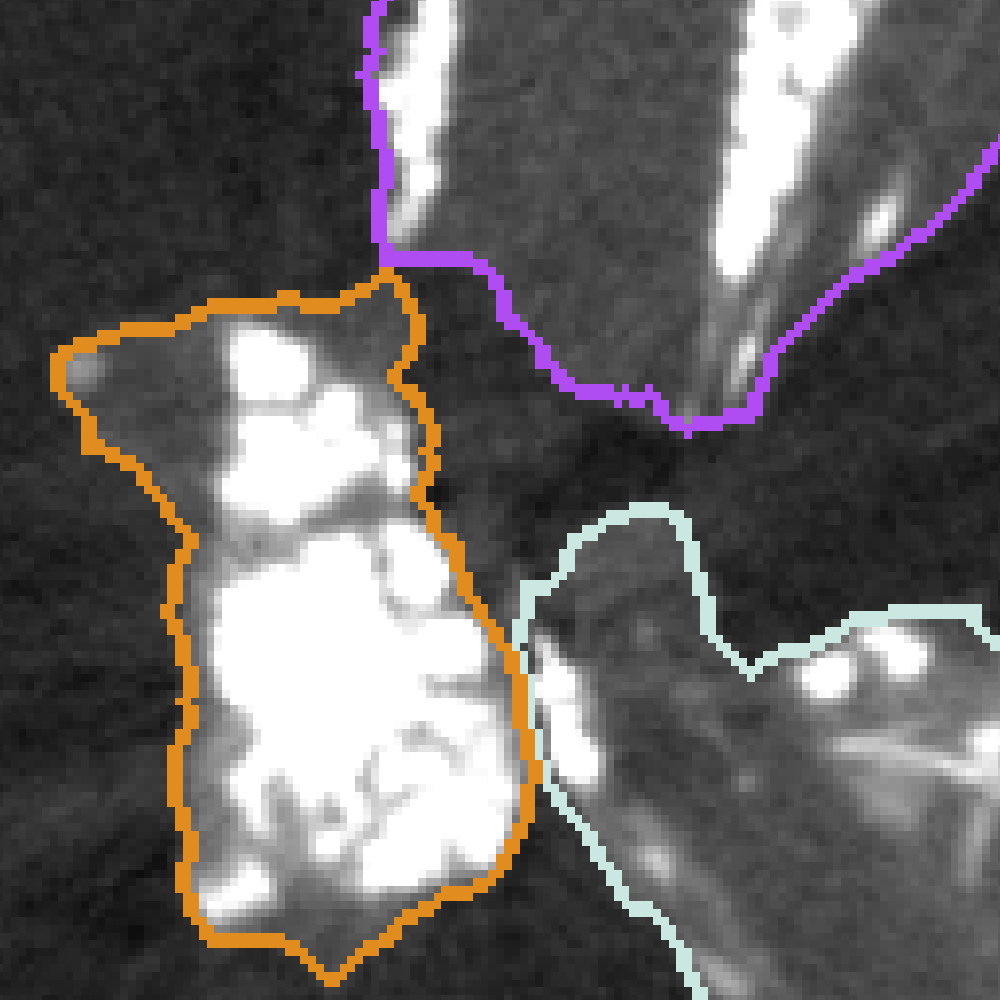}
                \caption{Split Particles}
        \end{subfigure}%
        \caption{Annotation process of separating touching particles in 3D to create an instance segmentation. (a) Touching particles are manually identified; (b) A marker is placed on each particle and a 2D border is drawn with our particle splitting tool; (c) The particles are successfully separated in 3D and each is assigned a unique identifier.}
        \label{fig:methodology:reference_annotation:blob_split}
\end{figure}

3D images of particle samples typically contain thousands of individual particles. Annotating entire images, even with the support of our proposed annotation pipeline, is too labor-intensive and would yield mostly redundant information for our method. Thus, utilizing the heterogeneity of the images, we take one or several representative smaller patches from each training sample for annotation. To enable fast and precise annotations the interactive 3D viewer Napari \cite{sofroniew2022napari} is used with multiple plugins to guide the annotation process. A guide for the annotation process is available at \href{https://github.com/MIC-DKFZ/ParticleSeg3D}{https://github.com/MIC-DKFZ/ParticleSeg3D}. \\
For each patch, a semantic segmentation of the particles is performed through the use of a Random Forest voxel classifier that is trained with scribble annotations. Random Forest classifiers operate on a broad set of handcrafted image features such as intensity gradients, image intensities, and gradient orientations making them more robust against CT imaging artifacts. The segmentation process is iteratively repeated with more refined scribbles until the result is satisfactory. Once completed, the remaining errors are corrected fully manually. An example of this random forest segmentation is shown in Figure \ref{fig:methodology:reference_annotation}.
The semantic segmentation is then converted into an instance segmentation by assigning all particles a unique identifier. However, particles touching each other cannot be separated in this step, and are incorrectly given only a single identifier, which must subsequently be corrected. To achieve this, we propose a custom particle-splitting tool. A marker is placed on each particle and a border is drawn between the particles. The particle-splitting tool then computes the geodesic distance \cite{wang2018deepigeos} of each voxel in both particles to the defined 2D border and separates them by applying a watershed algorithm on the geodesic distances with the particle markers as seeds for the algorithm. The result is a well-defined 3D border between the two particles in most cases requiring only a few seconds of annotation time per split. The process of this particle separation is depicted in Figure \ref{fig:methodology:reference_annotation:blob_split}.

\subsection{Training pipeline}\label{Training pipeline}

The training pipeline, depicted in Figure \ref{fig:methodology:training_and_inference_pipeline:train}, starts with a preprocessing stage to normalize the pixel intensities and homogenize the particle sizes to cope with the size changes of multiple orders of magnitude (Section \ref{Preprocessing}). Next, a border-core representation, shown to perform well for large-scale cell tracking \cite{isensee2021nnu} and thus suitable for identifying particles in the ten thousands, is used to map the reference instance segmentations to semantic segmentations (Section \ref{Border-core representation}). The border-core representations are then used for training nnU-Net, which has been proven highly successful on a multitude of diverse datasets (Section \ref{Model training}).

\subsubsection{Preprocessing}\label{Preprocessing}

Voxel intensities \(I\) of all images are normalized via z-scoring. To this end, the global mean intensity \(\mu\) and standard deviation of intensities \(\sigma\) are computed over all training images. Then, each voxel intensity is normalized \(I_{norm}\) with the following Equation \ref{fig:methodology:preprocessing:intensity_normalization}.

\begin{equation}
I_{norm} = \frac{I-\sigma}{\mu}
\label{fig:methodology:preprocessing:intensity_normalization}
\end{equation}

Particle sizes in our datasets vary substantially from 55 to 721 micrometers as determined through the equivalent particle diameter. At the same time, CT images are acquired at varying voxel spacings. Consequently, the size of particles in voxels varies substantially between images. Such a broad size distribution can hamper the training of the segmentation model, causing reduced performance and reduced robustness. To counteract this problem, we strive to homogenize the particle size distribution in our training data. To this end, the reference particle size of a sample is determined by measuring the diameter of a particle that is about average size within the sample. No sophisticated methods are required for this as the particle size only needs to be approximate. All training patches are then resized to a common average particle size.

\subsubsection{Border-core representation} \label{Border-core representation}

\begin{figure}[b]
\centering
        \begin{subfigure}[b]{0.3332\columnwidth}
                \centering
                \includegraphics[width=.95\linewidth]{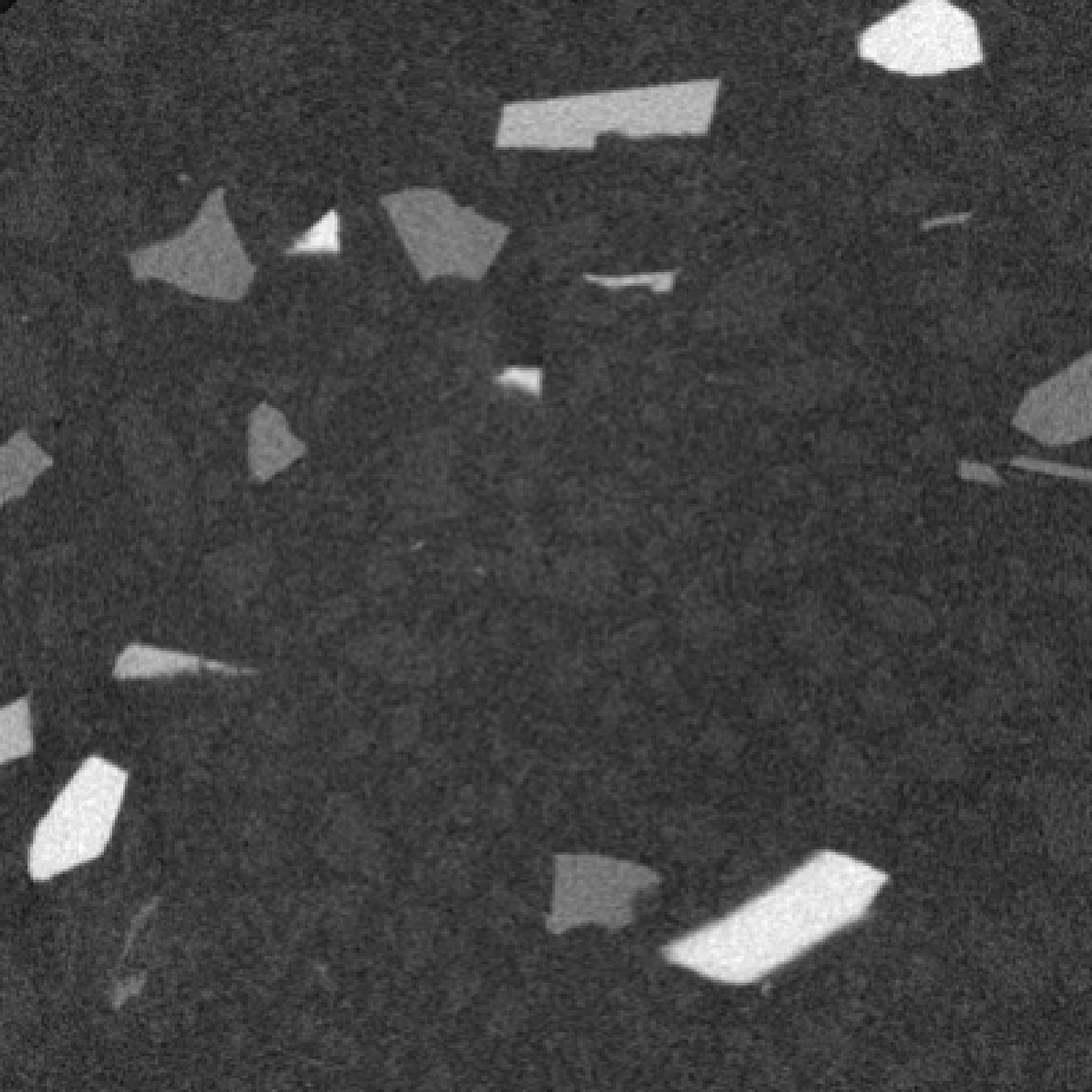}
                \caption{Patch}
        \end{subfigure}%
        \begin{subfigure}[b]{0.3332\columnwidth}
                \centering
                \includegraphics[width=.95\linewidth]{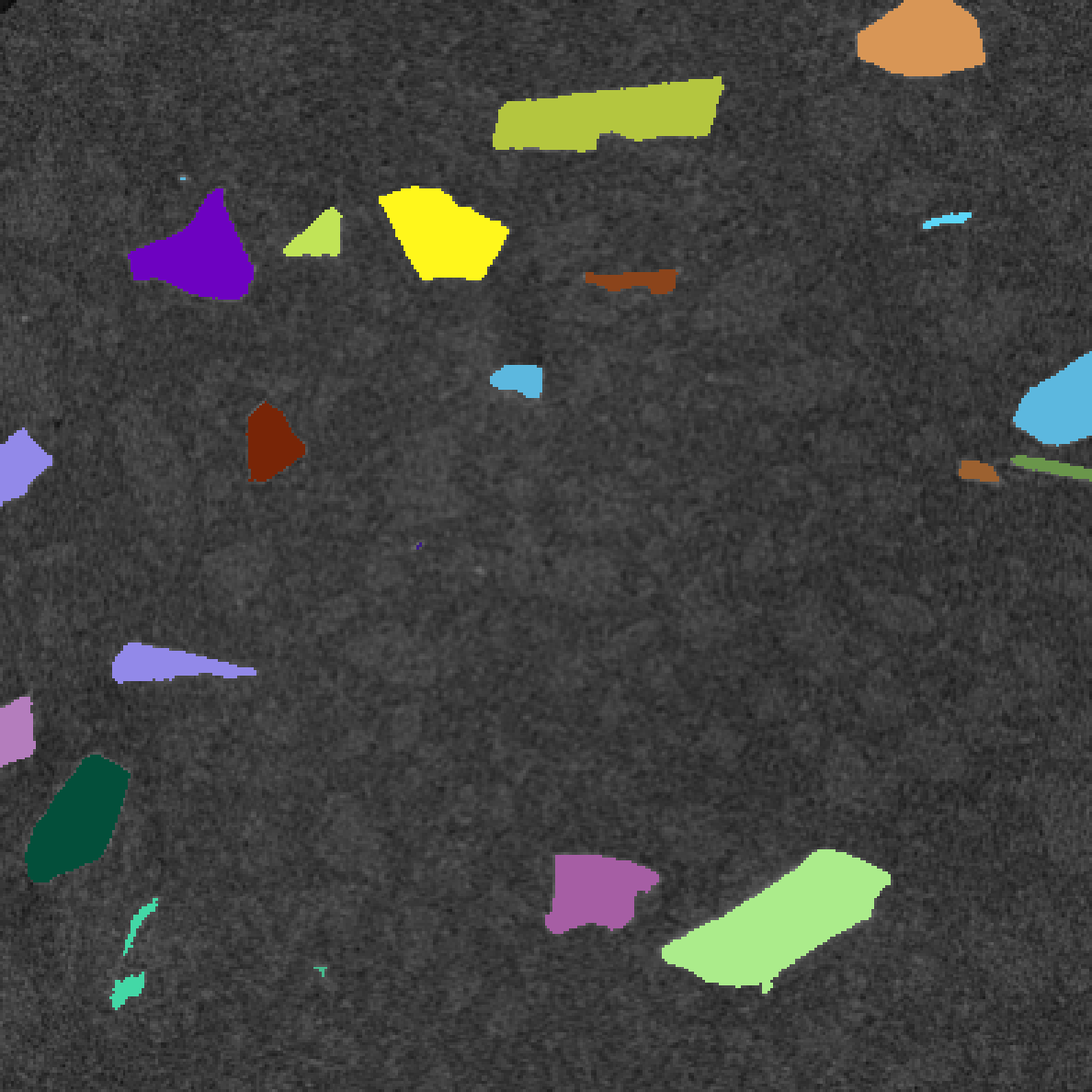}
                \caption{Instances}
        \end{subfigure}%
        \begin{subfigure}[b]{0.3332\columnwidth}
                \centering
                \includegraphics[width=.95\linewidth]{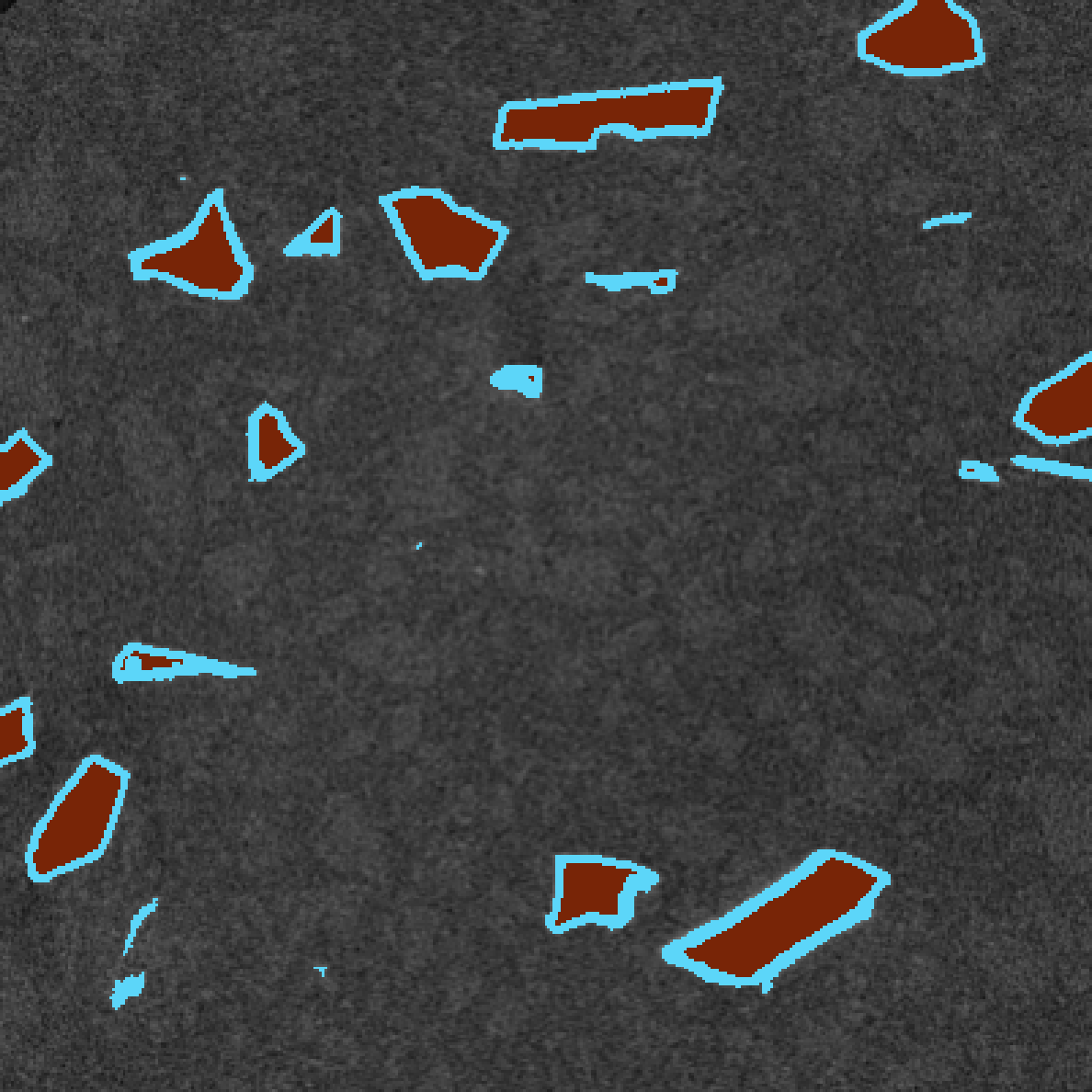}
                \caption{Border-Core}
        \end{subfigure}%
        \caption{The conversion process of a patch (a) from its instance segmentation (b) into its corresponding border-core representation (c).}
        \label{fig:methodology:border_core}
\end{figure}

A border-core representation enables any semantic segmentation model to be used for instance segmentation. In contrast to ad-hoc conversion methods such as watershed that often fail to correctly separate instances, a border-core representation is more advantageous due to the intrinsic properties of the representation that are subsequently learned by the model during training. An example of this representation is shown in Figure \ref{fig:methodology:border_core}. \\
Conversion to the border-core representation is achieved by eroding every instance with a fixed width in voxels (see section \ref{Training configuration}) referred to as border thickness to generate a core, while the eroded area represents the border of a particle. As a result, the intrinsic  properties of this representation are that every particle has the same border thickness and that no core of an instance is connected to a core of another instance. Training a model with such a representation enables the model to learn these properties and replicate them during inference. Such predicted border-core representation can then be easily converted back into instance segmentations by assigning each core a unique identifier and dilating the cores back to the original particle shape as defined by the border. \\
It is important to note that in some cases, a border-core prediction may produce multiple small cores for a single particle if the particle is thinner than the border thickness. Thus, the small extra cores are filtered away in an additional post-processing step to prevent miss-predictions. This small core removal filter computes the Euclidean distance of each core voxel to its border and subsequently removes any core if the percentage of distances smaller than a minimum distance is larger than a given threshold (see section \ref{Training configuration}).

\subsubsection{Model training} \label{Model training}
We use nnU-Net \cite{isensee2021nnu} as the semantic segmentation model in our training and inference pipeline. nnU-Net is a self-adapting 3D U-Net that creates a fingerprint of relevant properties from a dataset and automatically adapts important training hyperparameters accordingly. Originally developed for 3D biomedical CT and MRI data, the nnU-Net became a state-of-the-art model on multiple benchmark datasets and has also shown its efficacy in other domains \cite{grabowski2022self, li2022unsupervised, spronck2023nnunet, budelmann2023segmentation, altini2022fusion}. Its consistent performance across multiple domains makes it an ideal fit for the task of particle segmentation in high-resolution CT images. No modifications have been made to the nnU-Net training, except for stronger data augmentations, detailed in section \ref{Training configuration}, to better cope with the varying voxel intensity distributions of different materials, and a touching particle augmentation, described in the following.

\subsubsection{Touching particle augmentation}
To a large degree nnU-Net is able to learn on its own the intrinsic properties of a border-core representation based on the already existing touching particles in the training data. However, the number of touching particles in the training data used in our method is too small for nnU-Net to fully prevent the miss-prediction of touching particles during inference. This leads to some touching particles being merged into a single particle.
To reduce this error and to further improve the nnU-Net training, we introduce a touching particle augmentation. This augmentation is applied on every image patch in a batch during a training iteration with a certain probability. It selects a random particle within the current patch (particle P) and a random particle taken from the entire training dataset (particle D) and copies particle D next to particle P such that their particle edges touch each other. This way, nnU-Net learns over time a better understanding of how to correctly identify touching particles as separate instances and consequently predict a better border-core representation.

\subsection{Inference pipeline} \label{Inference pipeline}
Inference on a new image is performed by first normalizing its voxel intensities (Section \ref{Preprocessing}). This is followed by a sliding window prediction in which the image is divided into patches as the entire image is too large to store in GPU memory. For each patch, the particle size normalization (Section \ref{Preprocessing}) is performed on-the-fly and the patch is subsequently passed to the model for predicting the border-core segmentation (Section \ref{Border-core representation}). Once a patch is predicted, its local patch prediction is inserted into its original position in the global prediction of the sample, aggregating all patches during the inference process into a final prediction. This prediction is at last converted into an instance segmentation and resampled back to its original particle size, concluding the inference process. However, not all images can be processed in this fashion as the memory required for the predicted patches can be immense. Therefore, instead of the naive sliding window approach, we employ a novel chunk-patch-based sliding window approach proposed by us to solve this issue as described in Section \ref{Chunk sampling and aggregation}. A depiction of the complete inference pipeline is shown in Figure \ref{fig:methodology:training_and_inference_pipeline:inference}.

\subsubsection{Chunk sampling and aggregation} \label{Chunk sampling and aggregation}
\begin{figure}[b]
\centering
        \begin{subfigure}[b]{0.49\columnwidth}
                \centering
                \includegraphics[width=.95\linewidth]{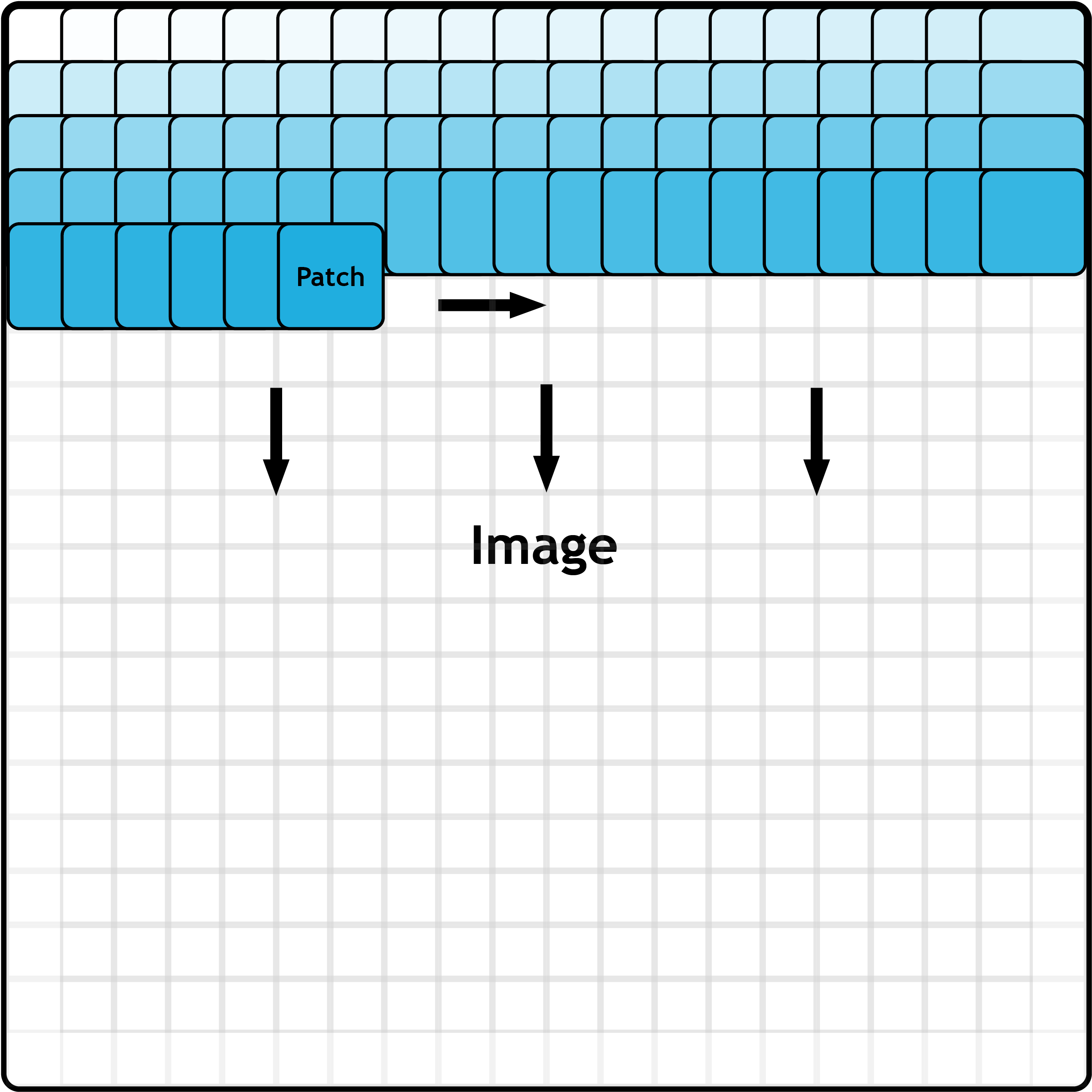}
                \caption{Patch Sampler}
                \label{fig:methodology:inference_pipeline:patch_sampler}
        \end{subfigure}
        \begin{subfigure}[b]{0.49\columnwidth}
                \centering
                \includegraphics[width=.95\linewidth]{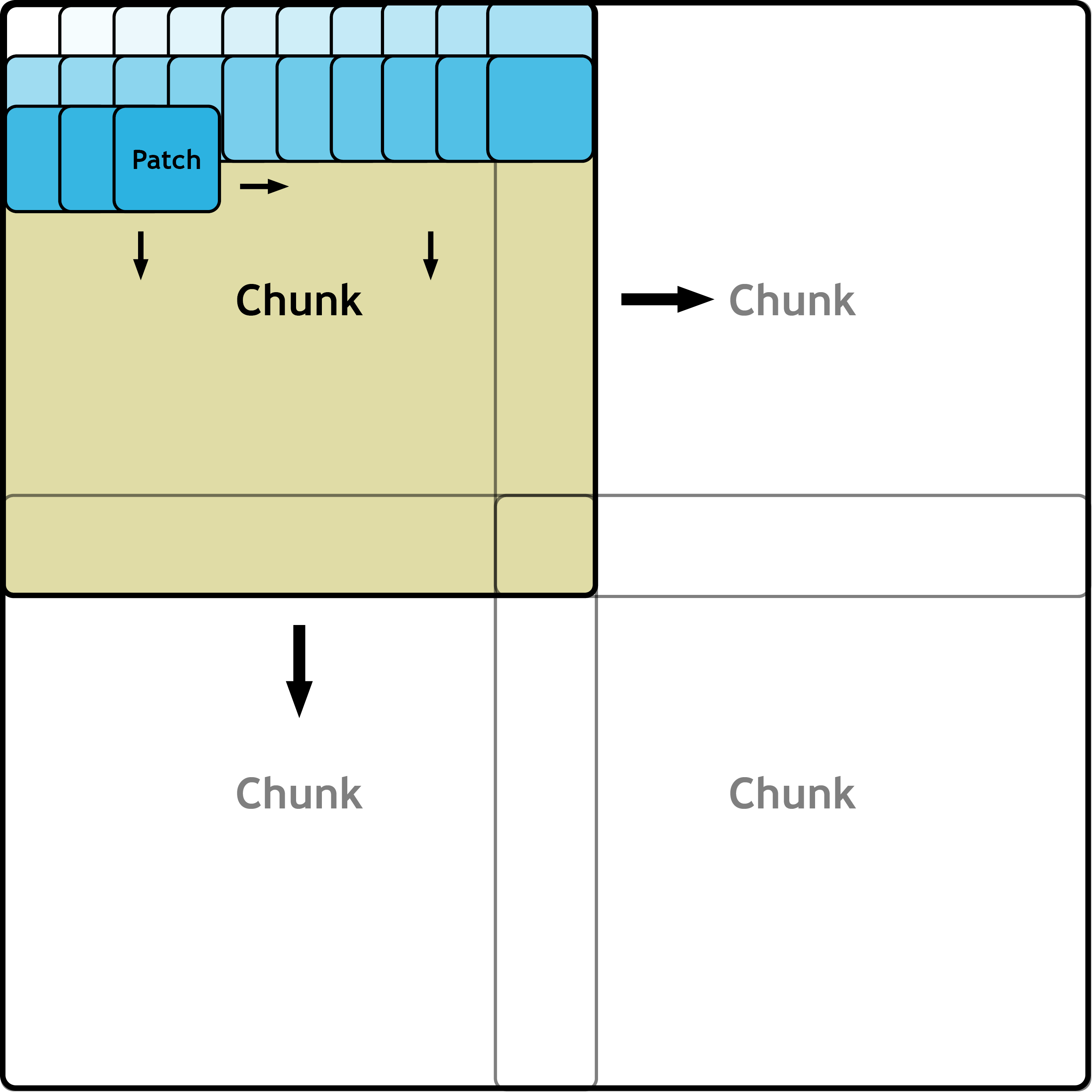}
                \caption{Chunk Sampler}
                \label{fig:methodology:inference_pipeline:chunk_sampler}
        \end{subfigure}
        \caption{Sampling process of a patch sampler compared to our developed chunk sampler that overcomes the limitations of the patch sampler.}
        \label{fig:methodology:inference_pipeline:sampler}
\end{figure}

A sliding window approach with a patch sampler as depicted in Figure \ref{fig:methodology:inference_pipeline:patch_sampler} is used during inference as the entire sample is too large to store in GPU memory. In order to still achieve the best possible accuracy, it has proven successful to use a patch overlap when using a sliding window approach. This means that except for the image edges, every voxel in the sample is predicted multiple times by the model, and the resulting predictions for a voxel are averaged. The model predicts class probabilities for every voxel in a patch and after the sliding window has reached the end of the sample, the predicted class probabilities of the sample are converted into the border-core segmentation. However, this approach reaches technical limitations for large images as in our case as the averaged patch predictions of every patch depend on all surrounding patches. Consequently, all predicted patches are required to be stored in memory, resulting in intractable memory consumption of often more than 100 GB. \\
To solve this issue, we developed a chunk-patch sampling strategy as depicted in Figure \ref{fig:methodology:inference_pipeline:chunk_sampler}. The sample is subdivided into multiple chunks. Each chunk has an overlap with the neighboring chunks of exactly one patch, ensuring that voxels at the edges of a chunk are also predicted multiple times through the patch overlap such that the prediction accuracy does not decrease at the chunk edges. A prediction is inferred for each chunk with the sliding window approach and subsequently saved to disk. This enables a memory-efficient inference of large images requiring not more than 5 GB of memory as only a single chunk is kept in memory at all times without compromising segmentation accuracy while adding only a minimal inference time overhead. 

\newpage

\section{Experimental Setup}

\subsection{Evaluation metrics} \label{Evaluation metrics}
The accuracy of the inferred instance segmentations is quantitatively measured with multiple metrics and can be divided into the categories of segmentation accuracy metrics (Section \ref{Segmentation quality metrics}) and separation accuracy metrics (Section \ref{Separation quality metrics}) which are introduced in the following sections.

\subsubsection{Segmentation accuracy metrics} \label{Segmentation quality metrics}
\newcommand\TP{\textit{TP}}
\newcommand\FP{\textit{FP}}
\newcommand\FN{\textit{FN}}
\newcommand\TN{\textit{TN}}
\newcommand\CM{\textit{CM}}
We use in total three metrics to measure the accuracy of the inferred predictions.
The \(F1\) score (Equation \ref{fig:experimental_setup_metrics:f1}), also known as \(Dice\) score, is a commonly used metric in image segmentation tasks to measure the ability of a method to correctly assign pixels to either foreground or background. However, it does not consider different particles and thus does not measure whether pixels are assigned to the correct particle. The \(F1_{match}\) score (Equation \ref{fig:experimental_setup_metrics:f1_match}) measures how well particles were recognized and separated, regardless of how accurately they are delineated. The \(F1_{instance}\) score (Equation \ref{fig:experimental_setup_metrics:f1_instance}) considers both segmentation accuracy as well as the correct identification of particles and is thus the metric most indicative of the performance of our method. \\
The ranges of all three metrics are within \([0,1]\) with a higher score indicating a better segmentation accuracy. All metrics are based on a confusion matrix \(\CM\) with true positives, false positives, true negatives, and false negatives being denoted as \(\TP\), \(\FP\), \(\TN\), and \(\FN\), respectively. Further, the confusion matrix \(\CM\) is computed in three different variations:
\begin{itemize}
  \item Let $\CM_{v}$ be the confusion matrix that is computed over all voxels \(v\) in a prediction and its respective reference.
  \item Let \(\CM_{m}\) be the confusion matrix that is computed over the count of all matched and mismatched particles \(m\) with a match being defined as two particles having a minimum 0.1 \(F1\) overlap of its particle voxels.
  \item Let \(\CM_{i}\) be the confusion matrix that is computed over the voxels \(i\) of a single particle instance in a prediction and its respective reference with \(I\) denoting all particle instances. Note that completely falsely predicted particles without ground truth and ground truth particles without any prediction are also included in the computation.
\end{itemize}

\begin{equation}
F1 = \frac{2\TP_{v}}{2\TP_{v}+\FP_{v}+\FN_{v}}
\label{fig:experimental_setup_metrics:f1}
\end{equation}

\begin{equation}
F1_{match} = \frac{2\TP_{m}}{2\TP_{m}+\FP_{ms}+\FN_{m}}
\label{fig:experimental_setup_metrics:f1_match}
\end{equation}

\begin{equation}
F1_{instance} = \frac{\sum_{i}^{I} \frac{2\TP_{i}}{2\TP_{i}+\FP_{i}+\FN_{i}}}{\sum_{i}^{I} 1}
\label{fig:experimental_setup_metrics:f1_instance}
\end{equation}

\subsubsection{Separation accuracy metrics} \label{Separation quality metrics}
\newcommand\GM{\textit{GM}}
\newcommand\GS{\textit{GS}}
\newcommand\MergerRatio{\textit{MergerRatio}}
\newcommand\MergerRatios{\textit{MergerRatios}}
\newcommand\SplitterRatio{\textit{SplitterRatio}}
When inferring an instance segmentation, a common challenge is an incorrect identification of touching particles as a single merged entity, referred to as \textit{Merger}, and conversely, the splitting of a single particle into multiple entities, referred to as \textit{Splitter}. The occurrence of such misidentifications should be minimized, as they can significantly impact downstream analysis. To assess the robustness of our approach to these errors, we introduce the $\MergerRatio$ and $\SplitterRatio$ metrics, which represent the percentage of total particles in a sample that are misidentified as \textit{Mergers} or \textit{Splitters}, respectively. \\
To define these metrics, we consider a set of all particles \(P\) and a set \(M\) of touching particles that are incorrectly merged into a single particle. We also define a set \(\GM\) of all touching particle sets \(M\) that are misidentified. The $\MergerRatio$ is then calculated as shown in Equation \ref{fig:experimental_setup_metrics:merger_ratio}. It is important to note that \(M\) and \(\GM\) are defined based on the predicted segmentation, while \(P\) is defined based on the reference segmentation. This enables fair comparisons between different methods, as the number of predicted particles can vary across methods.


\begin{equation}
\MergerRatio = \frac{(\sum_{M_{i}}^{\GM} |M_{i}|) - |\GM|}{|P|}
\label{fig:experimental_setup_metrics:merger_ratio}
\end{equation}


The $\SplitterRatio$ is defined similarly. We consider a set of all particles \(P\) and a set \(S\) of incorrectly predicted split particles, all part of the same reference particle. We also define a set \(\GS\) of all splitter sets \(S\).
The $\SplitterRatio$ is then obtained using the formula presented in Equation \ref{fig:experimental_setup_metrics:splitter_ratio}. As with the $\MergerRatio$, the definitions of \(S\) and \(\GS\) are based on the predicted segmentation, while \(P\) is defined based on the reference segmentation. By using this approach, we can compare the performance of different methods even if they predict different numbers of particles.
It is worth noting that both the $\MergerRatio$ and the $\SplitterRatio$ can theoretically range from 0 to infinity, but in practice, they are usually smaller than 1. A lower ratio indicates a lower number of \textit{Mergers} and \textit{Splitters} and thus better performance.

\begin{equation}
\SplitterRatio = \frac{(\sum_{S_{i}}^{\GS} |S_{i}|) - |\GS|}{|P|}
\label{fig:experimental_setup_metrics:splitter_ratio}
\end{equation}

\subsection{Training configuration} \label{Training configuration}
The training of the nnU-Net used in ParticleSeg3D is done in PyTorch with SGD optimizer, a learning rate of 1e-2, a weight decay of 3e-5, a momentum of 0.99, and 1000 epochs of training time. A target particle size of 60 voxels is used in the particle normalization stage. In the border-core to instance conversion stage, the small core removal filter uses a minimum distance of 1 with a threshold of 0.95. The parameters of the nnU-Net augmentations \textit{BrightnessTransform}, \textit{ContrastAugmentationTransform}, \textit{GammaTransform}, \textit{BrightnessGradientAdditiveTransform} and \textit{LocalGammaTransform} have been modified and are provided in our source code in detail. 

\subsection{Baselines} \label{Baselines}
We intend to compare our method against other frequently used methods quantitatively and qualitatively in section \ref{Results}. Consequently, we reviewed common segmentation and postprocessing strategies used to create instance segmentations \cite{hassan2012nondestructive, becker2016x, dominy2011characterisation, godinho2019volume} in order to derive the standardized instance segmentation method ThreshWater, which is described in the following section \ref{ThreshWater}. Further, we also compare against the predictions inferred from the \textit{Deep Learning Tool} of the commercial image processing software Dragonfly, which is described in more detail in section \ref{Dragonfly}.

\subsubsection{ThreshWater} \label{ThreshWater}
Individual particle characterization via instance segmentation is mostly performed through thresholding followed by a postprocessing step designed to label individual particles and separate the ones that are touching each other. We facilitate the same method and use it for comparison to our method. Considering that particle intensities differ greatly between samples as shown in Figure \ref{fig:datasets:intensities}, the threshold is manually tuned for each image. To combat poor performance in low-contrast images, we morphologically open the resulting noisy particle mask to suppress an overwhelming number of small false positive particle detections while retaining the integrity of the true positive larger particles. To convert this semantic segmentation into an instance segmentation where the particles are separated and can be processed individually, we again follow best practices and make use of a watershed-based \cite{van2014scikit} splitting procedure. Seed points are generated by eroding the particle mask. The seeds together with the masks are then used by the watershed algorithm to separate the particles. Design choices including the amount of morphological opening for the noise reduction and the amount of erosion for the core generation were systematically evaluated and optimized using the 41 train patches.

\subsubsection{Dragonfly} \label{Dragonfly}
Dragonfly (v2022.2.0.12227, Objects Research Systems, Montreal, Quebec, Canada) provides semantic segmentation models such as 2D U-Nets in its \textit{Deep Learning Tool}. 3D U-Nets are also available but at the time of writing cannot be trained due to an internally fixed patch overlap of NxNx1 voxels resulting in an infeasible large number of patches being sampled. Training of a 2D U-Net with default configuration as provided by Dragonfly was conducted on all slices of the 41 train patches. During inference, a semantic segmentation is predicted by Dragonfly, which we convert into an instance segmentation with the same procedure as in ThreshWater by eroding the particle mask to generate seed points for a subsequent watershed instance separation.

\section{Results \& Discussion} \label{Results}
Evaluation of ParticleSeg3D was conducted quantitatively (section \ref{Quantitative evaluation}), qualitatively (section \ref{Qualitative evaluation}), and with respect to inference time (section \ref{Inference time evaluation}) on an in-distribution and out-of-distribution test set which are elaborated in more detail in section \ref{Quantitative evaluation}. Reference segmentations for both test sets were created as described in section \ref{Reference annotation} and were used to evaluate our predictions. Further, we also evaluated the method ThreshWater (section \ref{ThreshWater}) and Dragonfly (section \ref{Dragonfly}) in the same fashion as our method and used both as baselines. Training of our method, ThreshWater and Dragonfly was exclusively carried out on the 41 train set patches as detailed in section \ref{Datasets}.

\subsection{Quantitative evaluation} \label{Quantitative evaluation}
Quantitative method performance was measured using the $F1$ score, the $F1_{match}$ score, and the $F1_{instance}$ score. In addition, we performed an in-depth analysis of the method’s ability to correctly split touching particles using the $\MergerRatio$ and the $\SplitterRatio$. For a complete description of the metrics used, see section \ref{Evaluation metrics}.
\subsubsection{In-Distribution evaluation} \label{In-Distribution evaluation}
\begin{figure}
    \centering
    \includegraphics[width=0.9\linewidth]{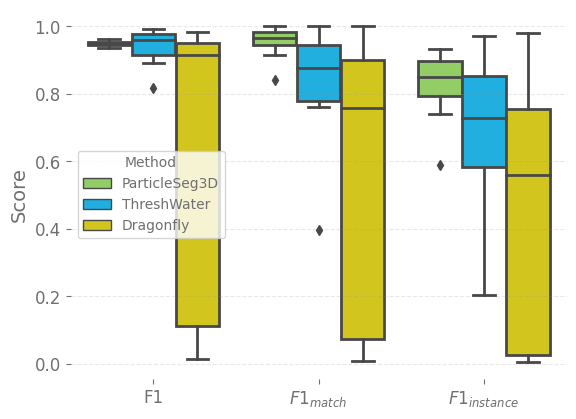}
    \caption{Segmentation quantification results on the in-distribution test set for each sample. Higher is better.}
    \label{fig:results:quantitative:in_distribution:seg}
\end{figure}

\begin{figure}
    \centering
    \includegraphics[width=0.9\linewidth]{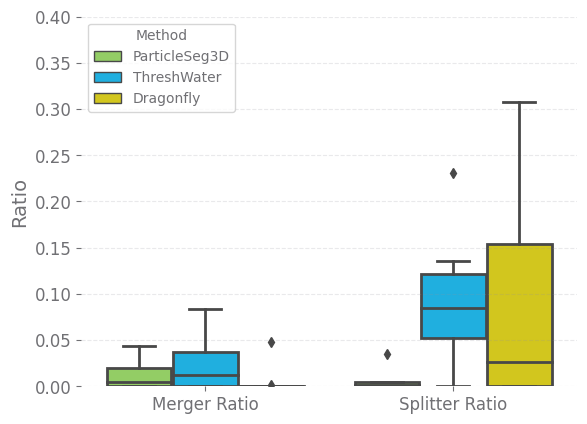}
    \caption{Separation quantification results on the in-distribution test set for each sample. Lower is better.}
    \label{fig:results:quantitative:in_distribution:sep}
\end{figure}

The in-distribution (ID) test set consists of samples with materials similar to the ones used during training and evaluation of such samples provides an estimate of the model's performance on types of samples the model will encounter in real-world applications. In total, the set contains 8 manually annotated patches from 8 samples as described in section \ref{Datasets}. We quantitatively evaluate the predictions of ParticleSeg3D as well as those of ThreshWater and Dragonfly using the metrics introduced in section \ref{Evaluation metrics}. However, we discuss the results for Dragonfly separately at the end of this section as Dragonfly performs significantly worse than the other methods. For ParticleSeg3D, ThreshWater and Dragonfly, the $F1$, the $F1_{match}$, and $F1_{instance}$ results are shown in Figure \ref{fig:results:quantitative:in_distribution:seg} and in Table \ref{table:results:quantitative:in_distribution:segmentation:comparison}. On average, a $F1$ of 94.84\% is achieved with ParticleSeg3D over all samples. Even on the individual samples, the $F1$ is above 93\%, indicating a high segmentation accuracy for every sample. Our average $F1$ is 1.01\% higher than that of ThreshWater, which is a substantial improvement at this level of precision. This is especially notable given the fact that ThreshWater uses a manually tuned threshold for each sample whereas our method is applied out-of-the-box. For the $F1_{match}$ score we achieve a score of 95.16\% on average over all samples, implying a high correct identification rate of individual particles as instances, which is amongst the most important properties of an instance segmentation. By comparison, ThreshWater only achieves an $F1_{match}$ score of 82.58\%, which is 12.58\% less than that of our method and indicates only a moderate identification rate of individual particles.
In terms of the $F1_{instance}$ metric, ParticleSeg3D achieves a score of 82.13\% on average while ThreshWater only achieves a score of 69.29\% and is thus 12.84\% worse compared to our method. It thus shows that ThreshWater performs poorly in comparison to our method in terms of individual particle segmentation and particle identification as measured by the $F1_{instance}$ metric. \\
\begin{table}
\caption{Segmentation quantification results on the in-distribution test set of ParticleSeg3D compared to ThreshWater and Dragonfly. Higher is better.}
\fontsize{7}{11}\selectfont
\begin{tabular}{| c | c | c | c |} 
\hline 
Name & $F1$ & $F1_{match}$ & $F1_{instance}$ \\ 
\hline 
Dragonfly & 0.6126 & 0.5587 & 0.4645 \\ 
\hline
ThreshWater  & 0.9383 & 0.8258 & 0.6929 \\ 
\hline
ParticleSeg3D  & 0.9484 & 0.9516 & 0.8213 \\ 
\hline
\end{tabular}
\label{table:results:quantitative:in_distribution:segmentation:comparison}
\end{table}
Following up on the segmentation accuracy evaluation, we continue with the discussion on the separation accuracy evaluation with the $\MergerRatio$ and the $\SplitterRatio$. Here, it is desirable to produce as few \textit{Mergers} and \textit{Splitters} as possible, which is often difficult as optimizing a method to produce fewer \textit{Mergers} can lead to more \textit{Splitters} and vice versa. A balanced tradeoff is therefore crucial. The results of ParticleSeg3D, ThreshWater and Dragonfly are depicted in Figure \ref{fig:results:quantitative:in_distribution:sep} and Table \ref{table:results:quantitative:in_distribution:separation:comparison}. The reference segmentations contain a total of 3020 particles. Of those 3020 particles only 198 particles are touching, which amounts to 7\% of the particles. This can be attributed to our sample preparation technique, in which a spacer is used to separate individual particles as much as possible from each other. Of those 198 touching particles, 64 \textit{Mergers} have been falsely created by ParticleSeg3D and 88 \textit{Mergers} by ThreshWater in the ID set, which results in a mean $\MergerRatio$ of 1.31\% and 2.67\% over the ID set, respectively. This means that on average ParticleSeg3D merges 1.31\% of particles with other particles compared to ThreshWaters 2.67\% in an ID sample. In terms of \textit{Splitters}, only 11 \textit{Splitters} in total are created by ParticleSeg3D in comparison to 329 \textit{Splitters} created by ThreshWater in the ID set, which results in a mean $\SplitterRatio$ of 0.58\% and 9.02\% over the ID set, respectively. This shows that the problem of \textit{Splitters} is successfully mitigated in ParticleSeg3D, while \textit{Mergers}, on the contrary, are still present to some extent, but reduced by more than half. ThreshWater performed similarly in terms of \textit{Mergers} but is prone to create a high number of \textit{Splitters}, which can have a severe impact on further downstream analysis tasks. The reasons for this frequent \textit{Splitter} generation are explored in section \ref{Splitters}. \\
To return to the results of Dragonfly, Dragonfly achieves a $F1$ of 61.26\%, a $F1_{match}$ of 55.87\% and a $F1_{instance}$ of 46.45\% on average, which is for all metrics significantly worse than both our method and that of ThreshWater. Regarding the $\MergerRatio$ and $\SplitterRatio$ of 0.62\% and 14.63\%, respectively, it seems that Dragonfly performs better in terms of \textit{Mergers} than ParticleSeg3D and ThreshWater. However, this is misleading as both the \textit{Mergers} and \textit{Splitters} cannot reliably be estimated on low accuracy predictions. Dragonfly's bad results indicates a lack of generalization already present on in-distribution data and can thus be expected to worsen on out-of-distribution data.
\begin{table}
\centering
\caption{Separation quantification results on the in-distribution test set of ParticleSeg3D compared to ThreshWater and Dragonfly. Lower is better. \phantom{g}}
\fontsize{7}{11}\selectfont
\begin{tabular}{| c | c | c | c | c |} 
\hline 
Name & Merger Ratio & Mergers & Splitter Ratio & Splitters \\ 
\hline 
Dragonfly & 0.0062 & 17 & 0.1463 & 84 \\ 
\hline
ThreshWater & 0.0267 & 88 & 0.0902 & 329 \\ 
\hline
ParticleSeg3D & 0.0131 & 64 & 0.0058 & 11 \\ 
\hline
\end{tabular}
\label{table:results:quantitative:in_distribution:separation:comparison}
\end{table}
\subsubsection{Out-Of-Distribution evaluation} \label{Out-Of-Distribution evaluation}

\begin{figure}[t]
    \centering
    \includegraphics[width=0.9\linewidth]{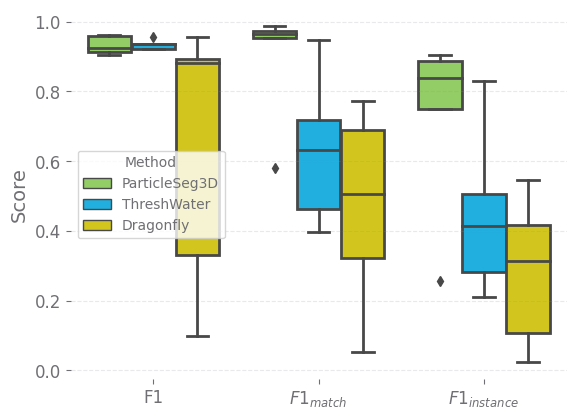}
    \caption{Segmentation quantification results on the out-of-distribution test set for each sample. Higher is better.}
    \label{fig:results:quantitative:out_of_distribution:seg}
\end{figure}

\begin{figure}
    \centering
    \includegraphics[width=0.9\linewidth]{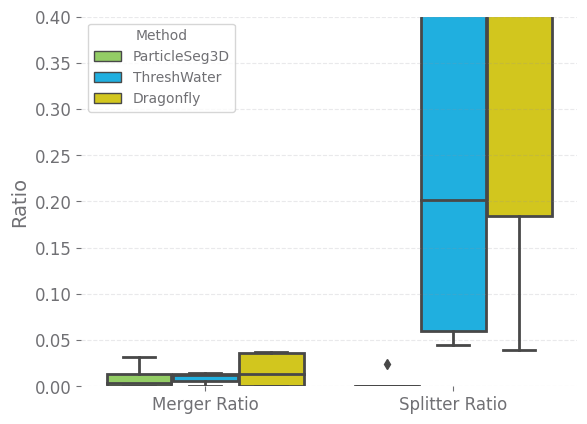}
    \caption{Separation quantification results on the out-of-distribution test set for each sample. Lower is better.}
    \label{fig:results:quantitative:out_of_distribution:sep}
\end{figure}

In contrast to the in-distribution (ID) test set, the out-of-distribution (OOD) test set contains samples with materials, particle sizes, and particle shapes that are not present in the training dataset. Thus, this evaluation provides valuable information on how well our method performs on samples with different characteristics not seen before and consequently highlights how it can be expected to generalize to a wider scope of use-cases. Our out-of-distribution test set contains 5 annotated patches from 5 different samples. The quantitative analysis is analogous to the previous section. The $F1$, the $F1_{match}$, and $F1_{instance}$ results of our method and ThreshWater are shown in Figure \ref{fig:results:quantitative:out_of_distribution:seg} and Table \ref{table:results:quantitative:out_of_distribution:segmentation:comparison}.
\begin{table}[t]
\centering
\caption{Segmentation quantification results on the out-of-distribution test set of ParticleSeg3D compared to ThreshWater and Dragonfly. Higher is better.}
\fontsize{7}{11}\selectfont
\begin{tabular}{| c | c | c | c |} 
\hline 
Name & $F1$ & $F1_{match}$ & $F1_{instance}$ \\ 
\hline 
Dragonfly & 0.6316 & 0.4686 & 0.2813 \\ 
\hline
ThreshWater  & 0.9347 & 0.6311 & 0.4488 \\ 
\hline
ParticleSeg3D  & 0.9327 & 0.8918 & 0.7277 \\ 
\hline
\end{tabular}
\label{table:results:quantitative:out_of_distribution:segmentation:comparison}
\end{table}
On average ParticleSeg3D reaches on the OOD set a $F1$ of 93.27\%, a $F1_{match}$ of 89.18\%, and a $F1_{instance}$ of 72.77\%. The OOD $F1$ and thus the general foreground prediction accuracy has only a slight reduction of 1.57\% when compared against our ID $F1$ score. By contrast, the $F1_{match}$ score has a reduction of 5.98\% on the OOD set. This dropoff between OOD and ID is even more considerable for the $F1_{instance}$ with a reduction of 9.36\%. \\
The reduction for both metrics can be traced to the sample Ore4-PS-High, which includes a high number of very small particles (<30$\mu$m) with only a diameter of 1-2 voxels that have very low contrast. Thus even for humans, these particles are almost indistinguishable from the background, making them hardly detectable for our model. We highlight the challenges associated with this sample as part of our qualitative evaluation in section \ref{Missing}. When Ore4-PS-High is excluded from the evaluation a $F1_{match}$ of 96.94\% and an $F1_{instance}$ of 84.52\% are achieved, which are on par with the results from the in-distribution set. This shows that ParticleSeg3D generalizes well to other materials, particle sizes, and particle shapes enabling the usage of our method on a wide scope of use-cases under the assumption that the particle diameter should be larger than 2 voxels on very low-contrast images. \\
The comparison of the OOD ThreshWater results to its ID results reveals that ThreshWater also achieves worse scores on the $F1_{match}$ and $F1_{instance}$ metrics with a reduction of 19.47\% and 24.41\%, respectively. However, this reduction originates not from the Ore4-PS-High sample but is the result of generally worse and highly varying performance on all samples in the OOD set. This originates mostly from badly segmented low-contrast particles and particles that have not been segmented at all by ThreshWater as we discuss in section \ref{Noisy} and \ref{Missing}, demonstrating the limitations of classical approaches with more complex samples. \\
Returning again to the results of Dragonfly, a $F1$ of 63.16\%, a $F1_{match}$ of 46.86\% and a $F1_{instance}$ of 28.13\% is achieved on average, which is as expected worse than on the ID set and shows Dragonfly's lack of generalization abilities in its training pipeline implementation.
We continue the out-of-distribution evaluation with the separation accuracy metrics $\MergerRatio$ and the $\SplitterRatio$ with the results shown in Figure \ref{fig:results:quantitative:out_of_distribution:sep} and Table \ref{table:results:quantitative:out_of_distribution:separation:comparison}. The reference segmentations consist of 1045 particles with only 29 particles (2.55\%) touching each other as a result of our sample preparation strategy. Of those 29 touching particles 14 \textit{Mergers} are falsely created by ParticleSeg3D in the OOD set with ThreshWater performing minimally better with only 10 \textit{Mergers}, which results in similar mean $\MergerRatios$ of 0.97\% and 0.91\% over the OOD set, respectively. However, when considering the number of \textit{Splitters}, ParticleSeg3D only produces 2 \textit{Splitters} in the OOD set, while ThreshWater produces 188, which is equal to a mean $\SplitterRatio$ of 0.48\% and 55.4\% over the OOD set, respectively. It can thus be concluded that our method is able to reliably identify individual particles as can be seen by its low $\MergerRatio$ and $\SplitterRatio$ even in samples with materials, particle sizes, and particle shapes unknown to our method. On the other hand, ThreshWater also achieves a low $\MergerRatio$, but demonstrates the same shortcomings as in the in-distribution evaluation with more than half of the detected particles being \textit{Splitters}.
\begin{table}
\caption{Separation quantification results on the out-of-distribution test set of ParticleSeg3D compared to ThreshWater and Dragonfly. Lower is better. \phantom{g}}
\fontsize{7}{11}\selectfont
\begin{tabular}{| c | c | c | c | c |} 
\hline 
Name & Merger Ratio & Mergers & Splitter Ratio & Splitters \\ 
\hline 
Dragonfly & 0.0172 & 23 & 0.3362 & 170 \\ 
\hline 
ThreshWater & 0.0091 & 10 & 0.554 & 188 \\ 
\hline
ParticleSeg3D & 0.0097 & 14 & 0.0048 & 2 \\ 
\hline
\end{tabular}
\label{table:results:quantitative:out_of_distribution:separation:comparison}
\end{table}
\subsection{Qualitative evaluation} \label{Qualitative evaluation}
\begin{figure}[h!]
\centering
        \begin{subfigure}[b]{0.95\columnwidth}
                \centering
                \caption*{Ore1-Comp3-Concentrate}
                \includegraphics[width=.32\linewidth]{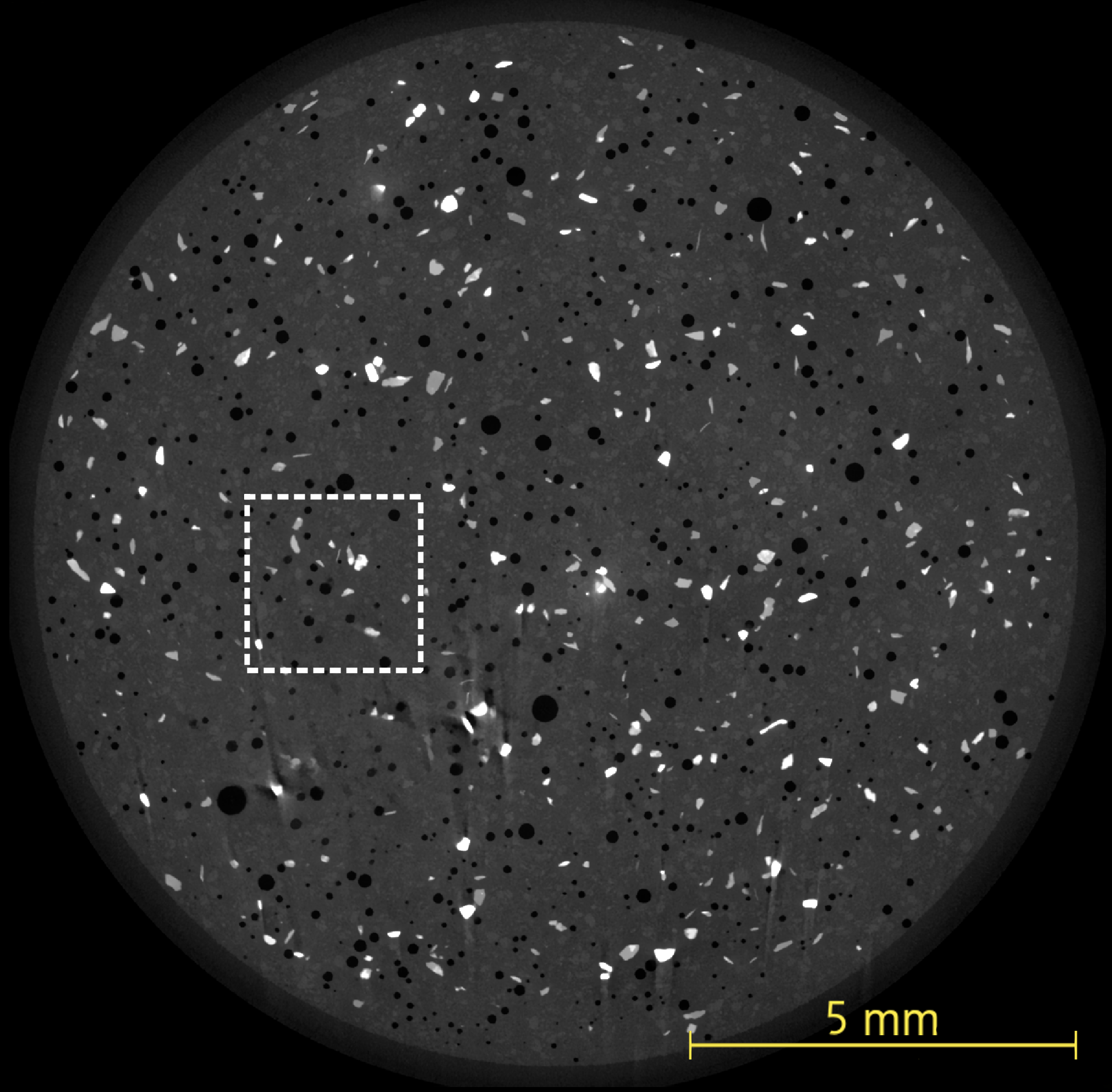}
                \hfill
                \includegraphics[width=.32\linewidth]{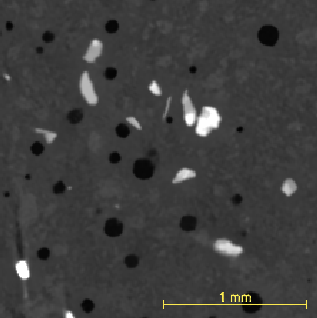}
                \hfill
                \includegraphics[width=.32\linewidth]{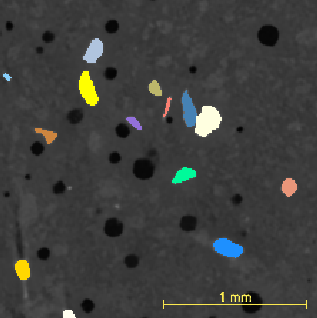}                
        \end{subfigure} \\[1ex]
        \begin{subfigure}[b]{0.95\columnwidth}
                \centering
                \caption*{Ore6}
                \includegraphics[width=.32\linewidth]{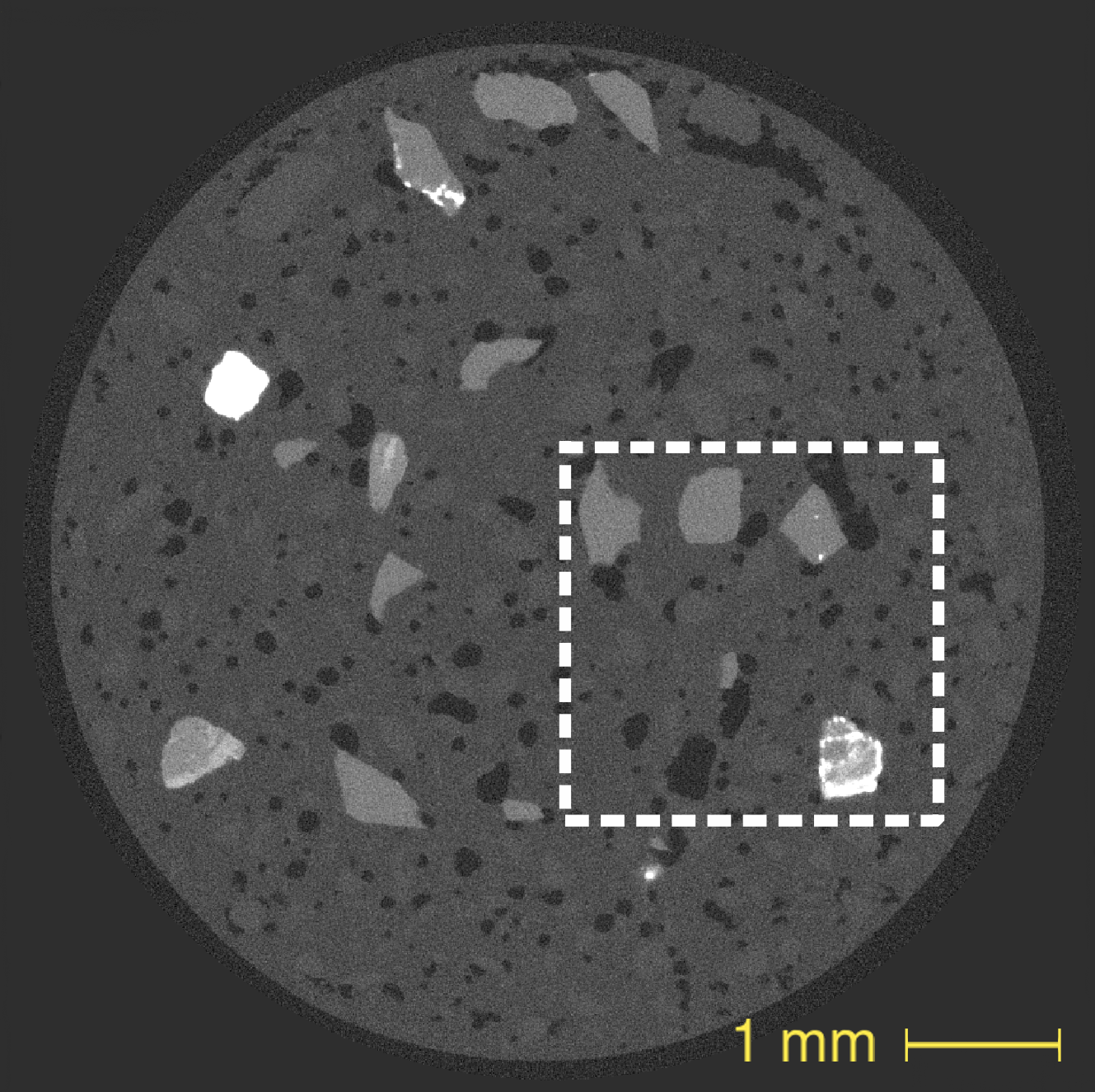}
                \hfill
                \includegraphics[width=.32\linewidth]{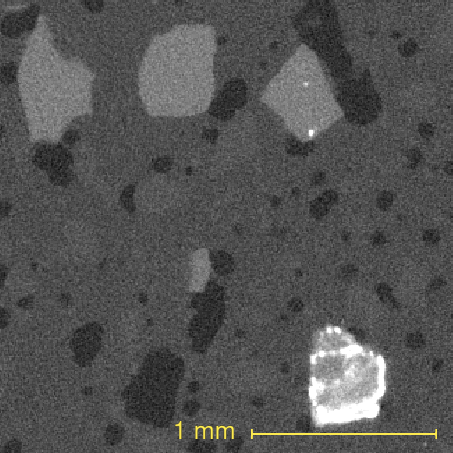}
                \hfill
                \includegraphics[width=.32\linewidth]{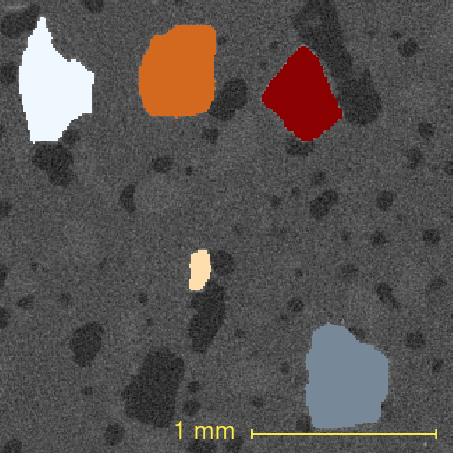} 
        \end{subfigure} \\[1ex]
        \begin{subfigure}[b]{0.95\columnwidth}
                \centering
                \caption*{Ore8}
                \includegraphics[width=.32\linewidth]{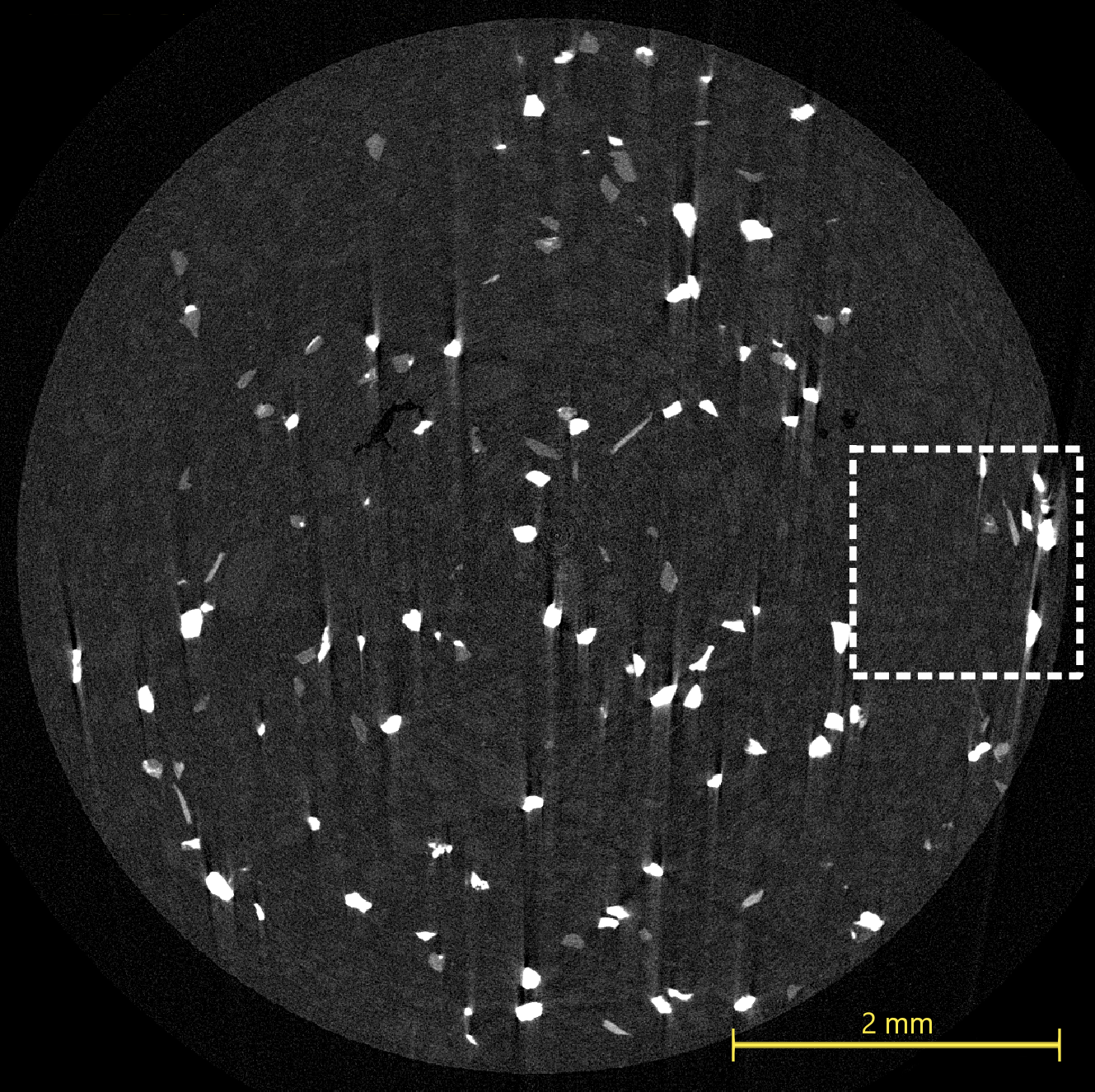}
                \hfill
                \includegraphics[width=.32\linewidth]{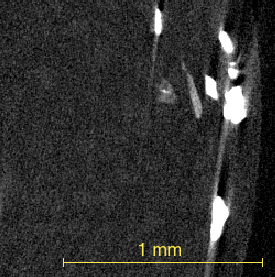}
                \hfill
                \includegraphics[width=.32\linewidth]{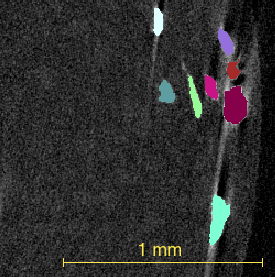}
        \end{subfigure}
        \caption{Overview of samples with different appearances, varying particle sizes, and CT imaging artifacts. Left: 2D view of the entire image; Middle: Zoomed crop of the image; Right: ParticleSeg3D prediction of the crop}
        \label{fig:results:qualitative:overview}
\end{figure}

We conducted a qualitative evaluation of the in-distribution (ID), out-of-distribution (OOD), and artifacts test samples with an overview of some samples and our inferred predictions given in Figure \ref{fig:results:qualitative:overview}. It can be seen already in the overview that ParticleSeg3D is able to segment all particles with high accuracy and is robust against varying particle sizes, shapes, and imaging artifacts.

\begin{figure*}[h!]
    \centering
    \includegraphics[width=1\textwidth]{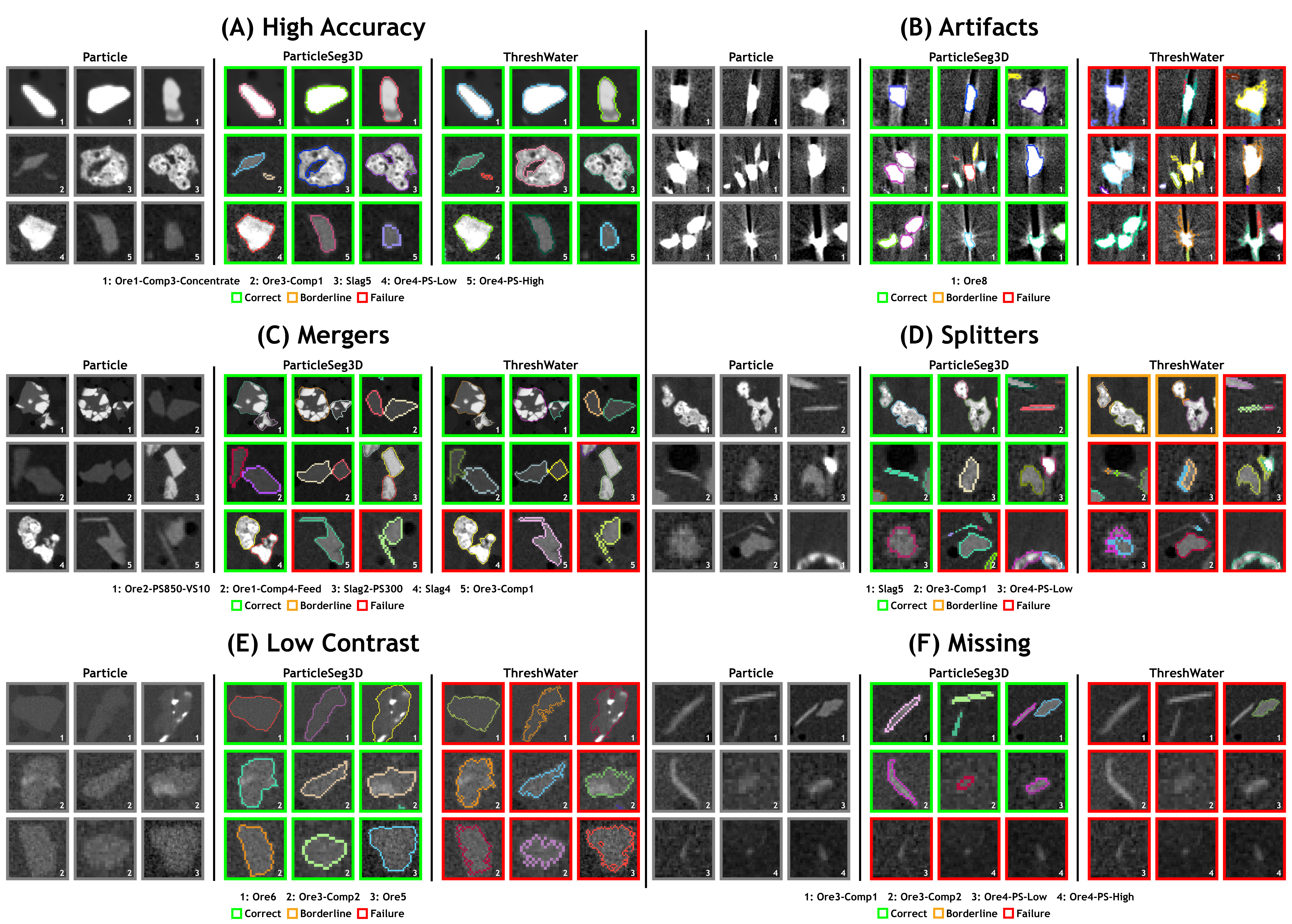} 
    \caption{Qualitative comparison of particles predicted by ParticleSeg3D and ThreshWater.}
    \label{fig:results:qualitative:all}
    \phantomlabel{.A}{fig:results:qualitative:all:high_quality}
    \phantomlabel{.B}{fig:results:qualitative:all:artifacts}
    \phantomlabel{.C}{fig:results:qualitative:all:mergers}
    \phantomlabel{.D}{fig:results:qualitative:all:splitters}
    \phantomlabel{.E}{fig:results:qualitative:all:low_contrast}
    \phantomlabel{.F}{fig:results:qualitative:all:missing}
\end{figure*}

Besides the general sample overview, several selected particles were inferred with ParticleSeg3D and ThreshWater for further discussion. The prediction of each of those particles can be categorized into one of six categories depending on the prediction accuracy and potential failure cases, which are discussed in the following subsections. The selected particles are depicted in Figure \ref{fig:results:qualitative:all} sorted per category with each particle prediction further marked as \textit{Correct (Green)}, \textit{Borderline (Orange)} or \textit{Failure (Red)} as determined through visual inspection. The low accuracy of Dragonfly's predictions is discussed separately in section \ref{Dragonfly qualitative evaluation}.

\subsubsection{High accuracy} \label{High accuracy}
ParticleSeg3D predicts most particles with high accuracy, as confirmed by our quantitative evaluation in section \ref{Quantitative evaluation}, while a majority are also accurately predicted by ThreshWater. A subset of particles from both the ID and OOD sets has been chosen for visualization in Figure \ref{fig:results:qualitative:all:high_quality}, demonstrating our method's generalization abilities to accurately delineate particle boundaries on diverse materials with a wide range of intensities, appearances, and even multi-phase particles.

\subsubsection{Artifacts} \label{Artifacts}
CT imaging artifacts such as streak artifacts caused by beam-hardening can significantly modify the particle's appearance and also affect neighboring particles, as illustrated in Figure \ref{fig:results:qualitative:all:artifacts}. These artifacts pose a particular challenge for classical segmentation methods. ThreshWater, for example, tends to segment entire streaks extending from particles, often leading to the merging of nearby particles. In contrast, our approach provides more realistic delineations of particle borders, even when multiple particles affected by artifacts are in close proximity. This highlights the robustness against such artifacts of deep learning techniques that leverage learned image features, as opposed to classical approaches that primarily rely on image intensities.

\subsubsection{Mergers} \label{Mergers}
In section \ref{Evaluation metrics}, a \textit{Merger} is defined as a group of two or more touching particles that are incorrectly identified as a single instance. Figure \ref{fig:results:qualitative:all:mergers} provides examples of merged particles and correctly separated particles. While ParticleSeg3D is successful in separating many touching particles, it sometimes fails when the particles are too intertwined or when one of the two touching particles is too thin. This is due to our small core filter as explained in section \ref{Border-core representation}, which removes the core of particles with a diameter of only 1-3 voxels, making it difficult to identify them as separate instances when they touch another particle. However, this situation is infrequent in practice, and our method produces a low number of \textit{Mergers}, as indicated in our quantitative evaluation in section \ref{Quantitative evaluation}. ThreshWater, on the other hand, produces noticeably more \textit{Mergers}, particularly when our method correctly identifies them as separate instances.

\subsubsection{Splitters} \label{Splitters}
In section \ref{Evaluation metrics}, a \textit{Splitter} is defined as a particle that is incorrectly identified as multiple particles. Figure \ref{fig:results:qualitative:all:splitters} presents examples of \textit{Splitter} particles, as well as correctly identified particles. ParticleSeg3D produces a negligible number of \textit{Splitters}, and no direct cause for these few cases could be identified. In contrast, ThreshWater produces a significantly higher number of \textit{Splitters}, particularly for low-contrast samples. This is because poor predictions on such samples can result in disconnected regions within a particle, leading to a \textit{Splitter}. Our method, on the other hand, is robust against these failure cases, as shown by our comparison results.

\subsubsection{Low contrast} \label{Noisy}
Low-contrast samples can pose a challenge in distinguishing between the background and particles, making it difficult to accurately predict particle borders. Figure \ref{fig:results:qualitative:all:low_contrast} provides examples of low-contrast particles. ThreshWater is highly affected by low-contrast particles and produces inconsistent particle borders in almost all cases. On the other hand, ParticleSeg3D is robust against such challenges and accurately predicts particle borders with consistency, even under these difficult low-contrast conditions.

\subsubsection{Missing} \label{Missing}
In some cases, the inference method may fail to detect a particle, resulting in no predicted particle mask. Figure \ref{fig:results:qualitative:all:missing} shows examples of such missed particles. ParticleSeg3D mostly misses particles with a diameter below 3 voxels in low-contrast, high-noise images, indicating that 3 voxels are the detection limit of our model. While smaller particles can be detected, their reliability is compromised. Hence, we propose using this detection limit as a guideline for future work. In contrast, ThreshWater has a worse detection limit and also fails on larger particles, in which the particle mask is filtered away due to the noise reduction step. The noise reduction step is necessary for ThreshWater, but leads to smaller particles being missed in the process.

\subsubsection{Dragonfly qualitative evaluation} \label{Dragonfly qualitative evaluation}
\begin{figure}[h!]
        \begin{subfigure}[b]{0.48\columnwidth}
                \centering
                \includegraphics[width=.95\linewidth]{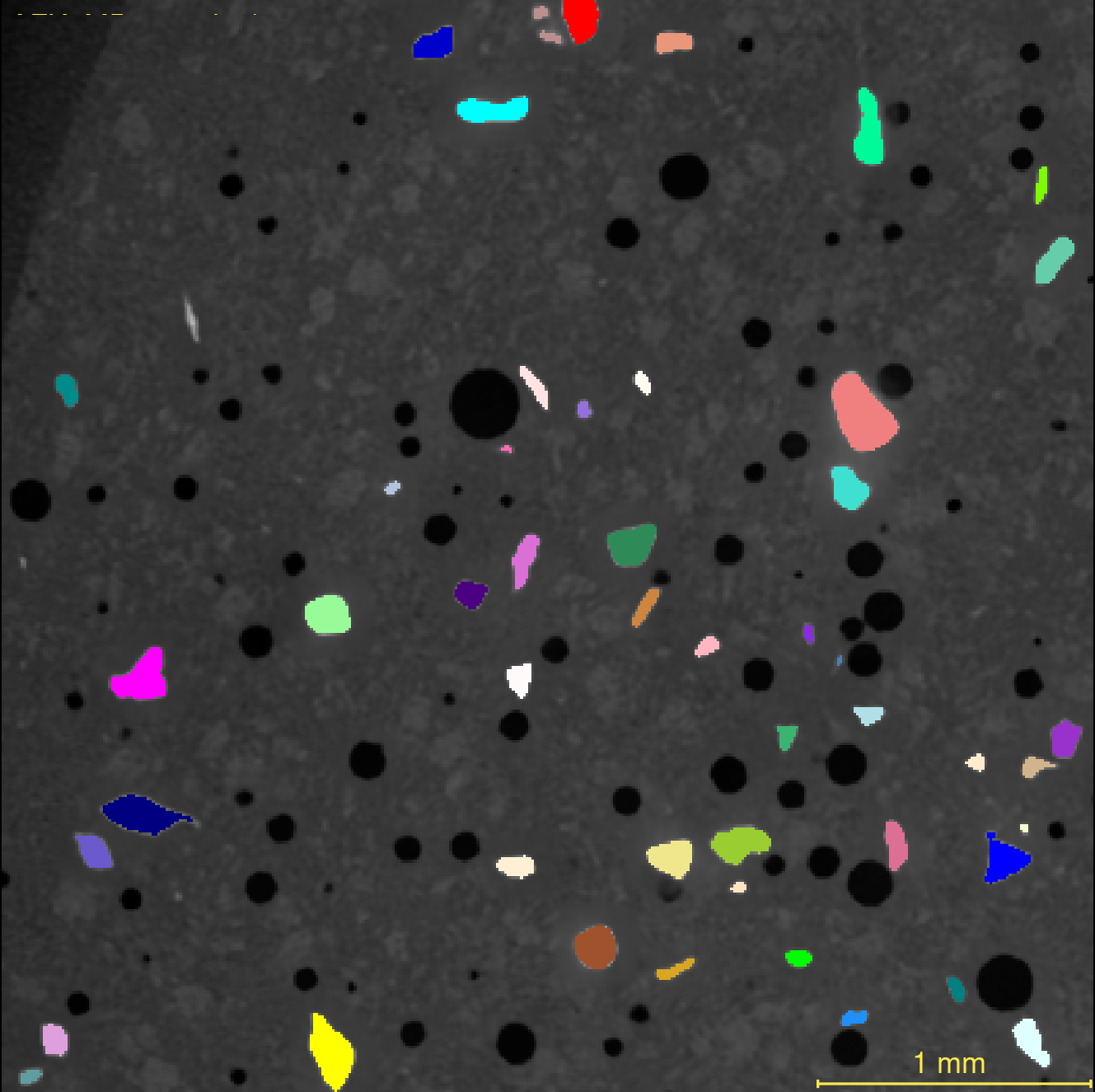}
                \caption{Ore1-Comp3-Concentrate} \label{fig:results:qualitative:dragonfly:a}
        \end{subfigure}%
        \begin{subfigure}[b]{0.48\columnwidth}
                \centering
                \includegraphics[width=.95\linewidth]{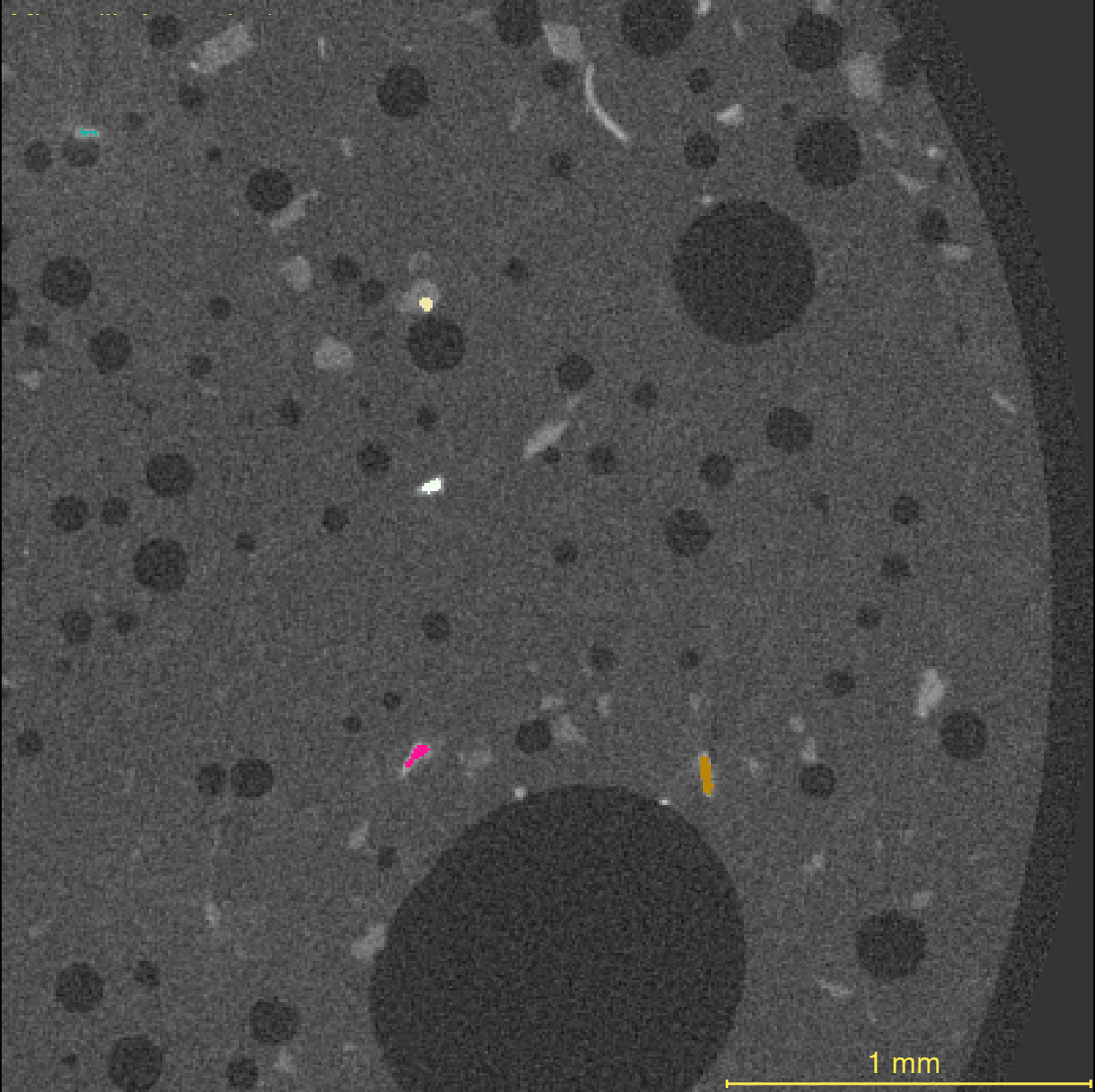}
                \caption{Ore3-Comp2} \label{fig:results:qualitative:dragonfly:b}
        \end{subfigure}  \\[1ex]
        \begin{subfigure}[b]{0.48\columnwidth}
                \centering
                \includegraphics[width=.95\linewidth]{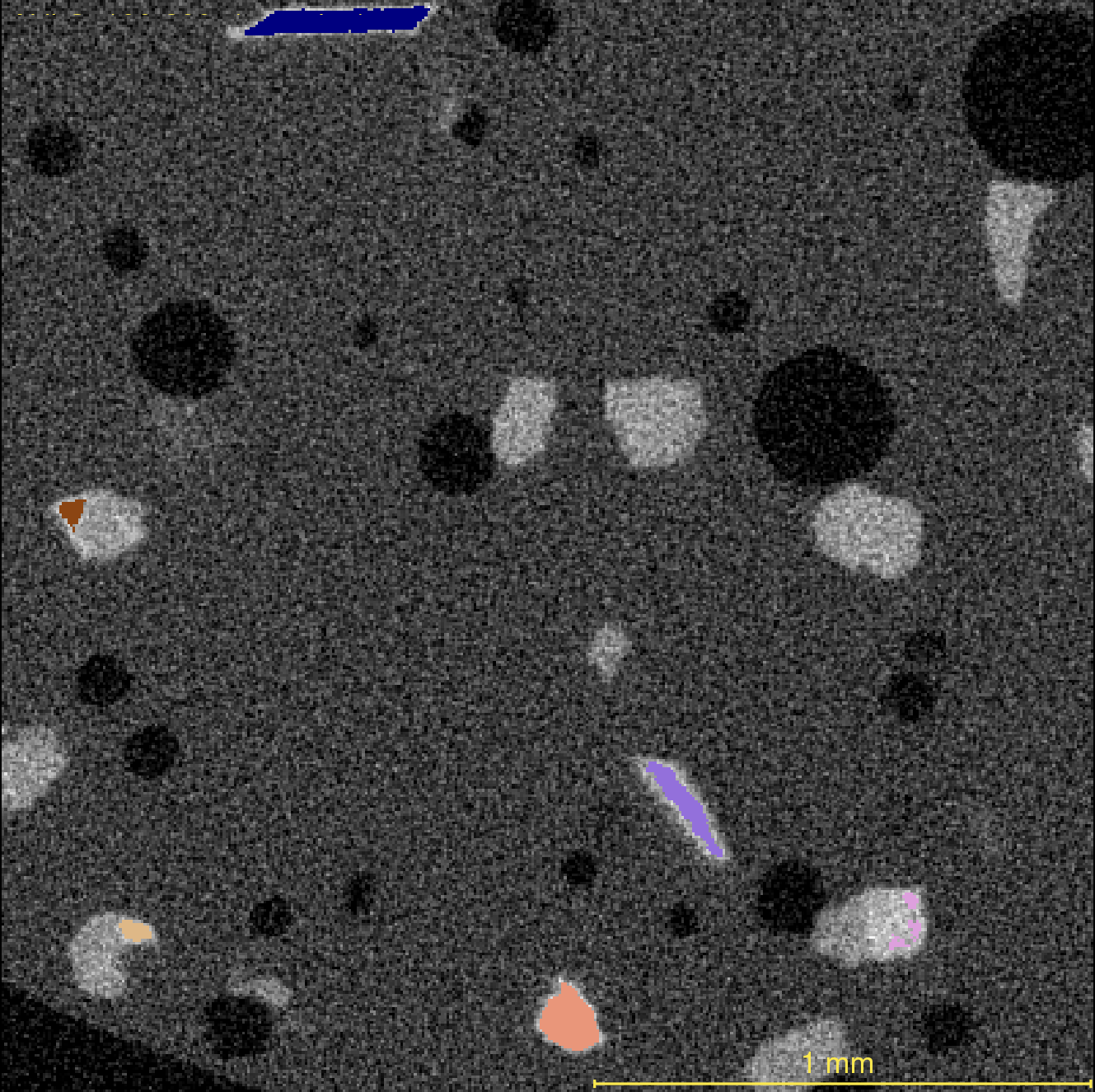}
                \caption{Ore5} \label{fig:results:qualitative:dragonfly:c}
        \end{subfigure}%
        \begin{subfigure}[b]{0.48\columnwidth}
                \centering
                \includegraphics[width=.95\linewidth]{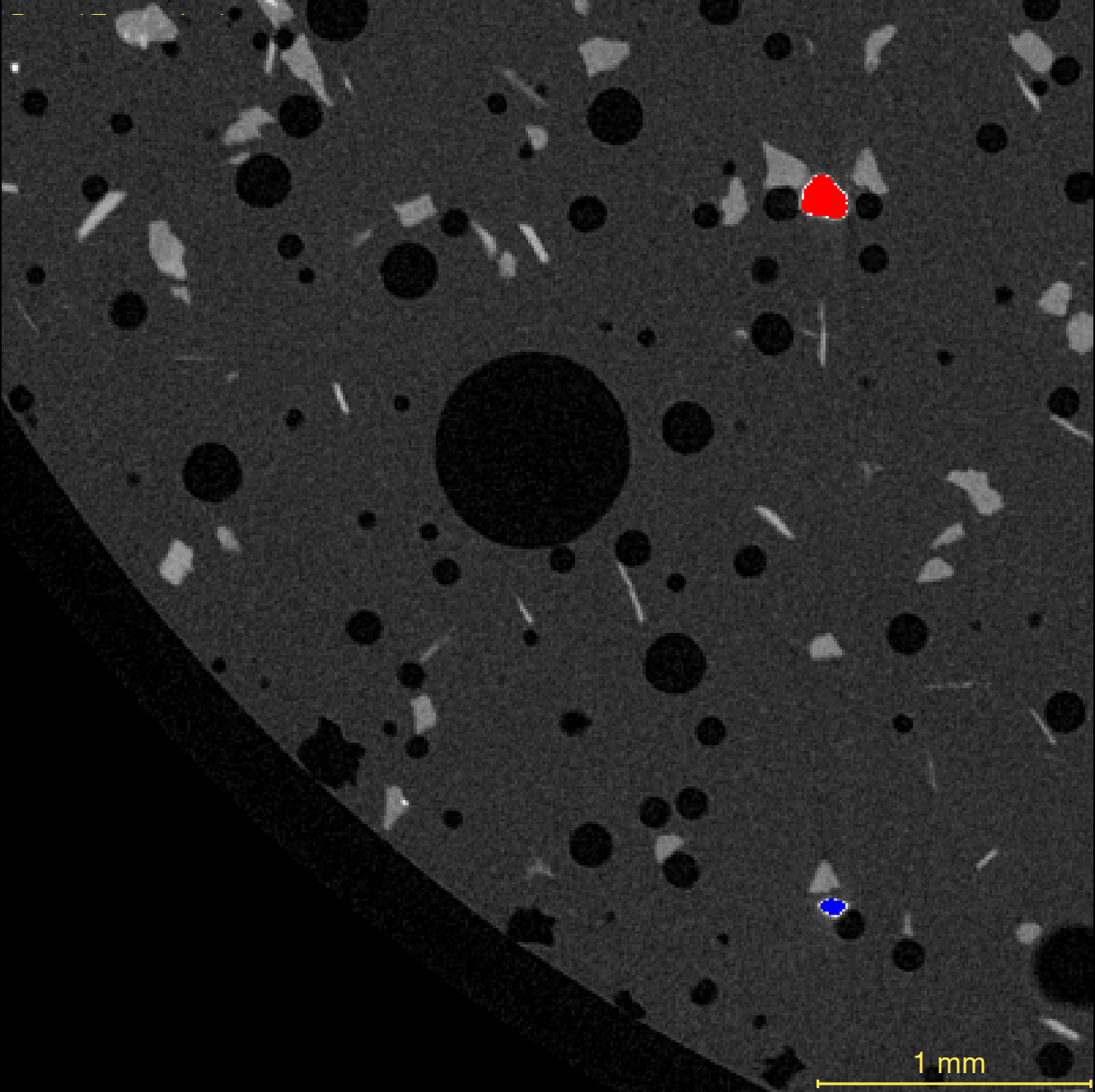}
                \caption{Ore1-Comp4-Feed} \label{fig:results:qualitative:dragonfly:d}
        \end{subfigure}%
        \caption{Qualitative examples of Dragonfly predictions. The accuracy of Dragonfly predictions can be reasonably good for some samples, as in (a), but is very limited for most other samples, as in (b), (c) and (d). Most particles are not identified as particles, and those that are, are predicted with significantly worse accuracy when compared to ParticleSeg3D or ThreshWater.}
        \label{fig:results:qualitative:dragonfly}
\end{figure}

We conducted a qualitative evaluation of the samples from the different tests with four sample prediction overviews given in Figure \ref{fig:results:qualitative:dragonfly}. The samples in Figure \ref{fig:results:qualitative:dragonfly:a} and Figure \ref{fig:results:qualitative:dragonfly:b} are part of the in-distribution (ID) set, while the samples in Figure \ref{fig:results:qualitative:dragonfly:c} and Figure \ref{fig:results:qualitative:dragonfly:d} are part of the out-of-distribution (OOD) set. Our analysis reveals that Dragonfly is capable of generating accurate predictions on the ID set as evidenced in Figure \ref{fig:results:qualitative:dragonfly:a}, but can also fail by almost not recognizing any particles as in Figure \ref{fig:results:qualitative:dragonfly:b}, suggesting its poor generalization already on ID data. This lack of generalization from Dragonfly is further exacerbated on the OOD set, by consistently not recognizing the majority of particles, exhibiting its inability to capture the underlying patterns of particle shapes, sizes, and intensity characteristics necessary to generalize to new, unseen images.

\subsection{Inference time evaluation} \label{Inference time evaluation}
\begin{figure}
        \begin{subfigure}[b]{1\columnwidth}
                \centering
                \includegraphics[width=.95\linewidth]{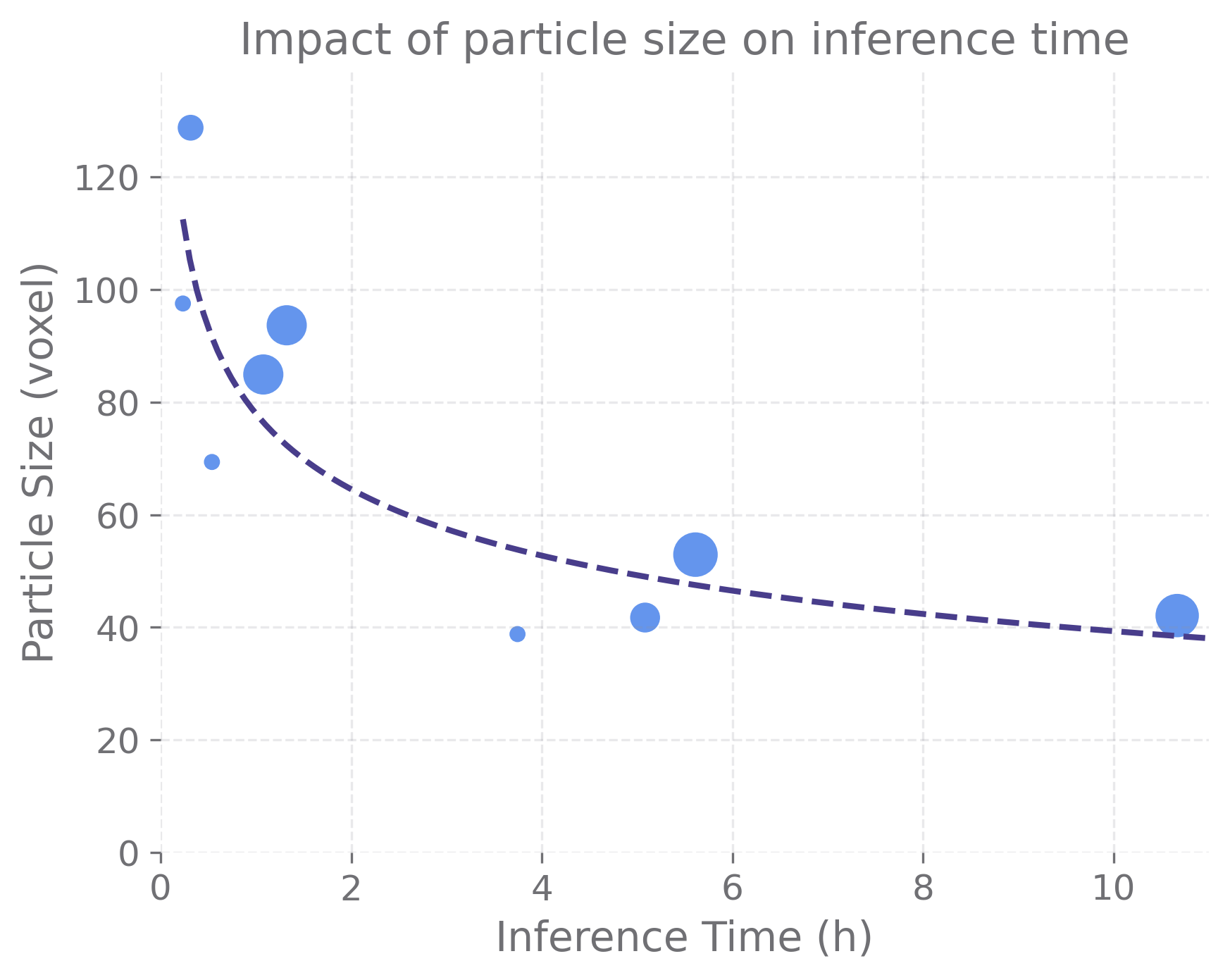}
        \end{subfigure}%
        
        \caption{Impact of the voxel particle size on the inference time. Evaluated on in-distribution and out-of-distribution test sets with the marker size being proportional to the respective image size. Expected inference time per voxel particle size is represented as a dashed interpolated function.}
        \label{fig:results:inference_time:particle_size}
\end{figure}

We conduct an evaluation on the impact of different particle and image sizes on the inference time and deduce advantages and limitations of ParticleSeg3D. Both the in-distribution and out-of-distribution test sets are used for the analysis. Given some fixed hardware, here a Nvidia A100 40GM PCIe GPU (Nvidia, Santa Clara, USA), the inference time only depends on the number of patches extracted and passed to the nnU-Net for a given image (Section \ref{Inference pipeline}). The number of patches, in turn, depends on the image size and the $ParticleSize$ in voxels, with the latter determined manually (see Section \ref{Preprocessing}) and computed with Equation \ref{equation:results:inference_time}.
\begin{equation}
\ParticleSizeVoxel = \frac{\ParticleSizeMM}{\VoxelSpacingMM}
\label{equation:results:inference_time}
\end{equation}
Figure \ref{fig:results:inference_time:particle_size} shows the impact of the different particle sizes on the inference time, with different image sizes being represented as circles with varying diameters. On average, our dataset has a particle size of 60±32 voxels and an image size of (997±287, 1256±218, 1256±218) voxels. The inference time for one image is in the median 4.41 hours. When inspecting Figure 5.3.D, we see that with a particle size of $\sim$60 the inference duration is $\sim$2 hours and with a particle size of $\sim$120 this duration is reduced to only $\sim$20 minutes. Conversely, with a particle size of $\sim$40, the inference time quickly increases to $\sim$10 hours. This exponential decline and rise in inference times originates from the particle size normalization (Section \ref{Preprocessing}) as particles need to be rescaled resulting in much larger or smaller effective image sizes. While having a manageable impact down to a particle size of $\sim$50 with $\sim$5 hours inference time, our method becomes significantly slower for smaller particle sizes.
\begin{equation}
\ParticleSizeMM = \ParticleSizeVoxel * \VoxelSpacingMM
\label{equation:results:inference_time2}
\end{equation}
Thus ParticleSeg3D provides fast inference times with appropriate voxel particles sizes, while still being manageable down to a voxel particle size of 50. Compared to the current alternative deep learning methods that require a multi-day retraining on each sample due to their lack of generalization, ParticleSeg3D provides a faster and more accurate individual particle characterization.
Further, with the goal of using the model for a large number of samples, we focused on striking a good balance between segmentation accuracy and inference times. For narrower application windows, a specifically optimized target particle size for the particle size normalization along with a retraining of our model could be conceivable to improve inference times and thus reduce the time for an individual particle characterization even further. 

\section{Conclusion}
Minerals, metals, and plastics are finite and must be acquired cost-effectively and responsibly through advances in processing and recycling. Hereby, the characterization of individual particles plays a crucial role in enabling and optimizing automation in the face of rising demand. In the wake of increasingly sophisticated particle processing techniques, there is an ever-growing need for characterizing individual particles over bulk material characterization. Throughput and the ability to quickly analyze new, unique samples is key to catalyze future progress in particle processing technology. \\
One key step towards characterizing individual particles is to isolate them from 3D images containing up to tens of thousands of such particles through a process called instance segmentation. In this manuscript, we presented ParticleSeg3D, a new deep learning framework capable of doing just that. By leveraging the power of the nnU-Net framework, as well as the border-core representation to perform instance segmentation, ParticleSeg3D is able to substantially outperform the currently used approaches in the field with respect to segmentation accuracy, its ability to separate touching particles as well as its robustness to varying particle sizes, imaging artifacts, and unseen materials. Our extensive evaluation of the out-of-distribution test set underlines how our method can be applied even to previously unseen materials without requiring retraining. Contrary to traditional methods, where new training data needs to be collected and the algorithm needs to be retrained for each image, our solution constitutes a substantial workflow improvement and unlocks previously inconceivable levels of scalability. For the vast majority of samples, it can simply be used to make predictions with no further human input required. \\
ParticleSeg3D is thus well positioned to greatly enhance scalability, precision, and automation of individual particle 3D characterization in a standardized way, and will thus act as a catalyst for accelerating the development of processing, and recycling strategies. All code and datasets are made publicly available along with instructions on how to use them.

\section*{Acknowledgement}
Part of this work was funded by Helmholtz Imaging (HI), a platform of the Helmholtz Incubator on Information and Data Science. This work has received financing from the Deutsche Forschungsgemeinschaft within SPP 2315, project 470202518.

\section*{Conflict of Interest}
The authors declare that they have no known competing financial interests or personal relationships that could have appeared to influence the work reported in this paper.




\bibliographystyle{cas-model2-names}

{\normalfont\bibliography{cas-refs}}





\end{document}